\pdfoutput=1

\documentclass[11pt]{article}
\usepackage{placeins}
\usepackage[preprint]{acl}

\usepackage{times}
\usepackage{latexsym}

\usepackage[T1]{fontenc}

\usepackage[utf8]{inputenc}

\usepackage{microtype}

\usepackage{inconsolata}

\usepackage{graphicx}
\usepackage{amssymb}
\usepackage{enumitem}
\usepackage{amsmath}
\usepackage{colortbl}
\definecolor{lightgray}{gray}{0.9}
\definecolor{deepgreen}{rgb}{0,0.5,0}
\usepackage{multirow}
\usepackage{amsfonts}
\usepackage{booktabs}
\usepackage{algorithm}
\usepackage{algorithmic}
\usepackage{supertabular}
\usepackage{longtable}
\usepackage{tabu} 
\usepackage{ragged2e} 
\usepackage{makecell}

\title{
\textbf{$A^3$}-Bench: Benchmarking Memory-Driven Scientific Reasoning \\via \underline{A}nchor and \underline{A}ttractor \underline{A}ctivation}

\author{Jian Zhang$^{1}$, Yu He$^{1}$, Zhiyuan Wang$^{1}$, Zhangqi Wang$^{1}$, Kai He$^{2}$,\\
\bf{Fangzhi Xu$^{1}$, Qika Lin$^{2}$, Jun Liu$^{1}$\thanks{Corresponding author}}\\
	$^{1}$Xi'an Jiaotong University\;
    $^{2}$National University of Singapore\\
    \texttt{zhangjian062422@stu.xjtu.edu.cn}, \texttt{liukeen@xjtu.edu.cn}}

\begin{document}
\maketitle

\begin{abstract}

Scientific reasoning relies not only on logical inference but also on activating prior knowledge and experiential structures. Memory can efficiently reuse knowledge and enhance reasoning consistency and stability. However, existing benchmarks mainly evaluate final answers or step-by-step coherence, overlooking the \textit{memory-driven} mechanisms that underlie human reasoning, which involves activating anchors and attractors, then integrating them into multi-step inference.
To address this gap, we propose \textbf{$A^3$-Bench}~\footnote{\url{https://a3-bench.github.io/}}, a benchmark to evaluate scientific reasoning through dual-scale memory-driven activation, grounded in \underline{A}nchor and \underline{A}ttractor \underline{A}ctivation. 
First, we annotate 2,198 science reasoning problems across domains using the SAPM process(\underline{s}ubject, \underline{a}nchor \& attractor, \underline{p}roblem, and \underline{m}emory developing). Second, we introduce a dual-scale memory evaluation framework utilizing anchors and attractors, along with the \textit{AAUI} (\underline{A}nchor--\underline{A}ttractor \underline{U}tilization \underline{I}ndex) metric to measure memory activation rates. Finally, through experiments with various base models and paradigms, we validate $A^3$-Bench and analyze how memory activation impacts reasoning performance, providing insights into memory-driven scientific reasoning.

\end{abstract}

\section{Introduction}

\begin{figure}[t]
 \includegraphics[width=\columnwidth]{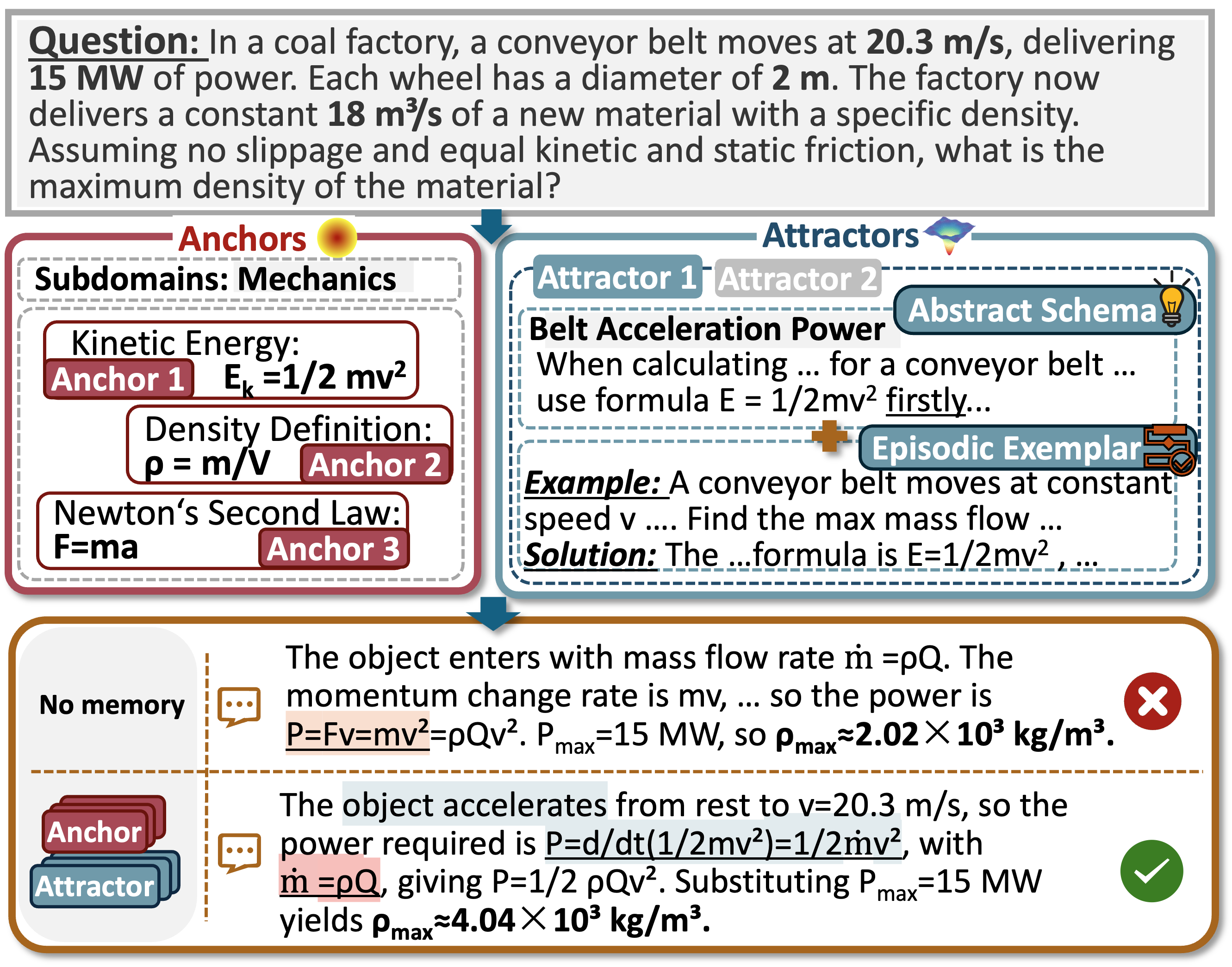}
 \caption{Comparison of reasoning paths on OlympiadBench. Activating anchors and attractors corrects the derivation path relative to no memory.}
 \label{fig_example}
\end{figure}

Scientific reasoning tasks~\citep{zhang2026maps,zhang2025physreason}, covering disciplines like math, physics, and chemistry, are essential for evaluating the ability of Large Language Models (LLMs) to integrate complex cognitive operations. Unlike traditional language tasks, scientific reasoning requires not only knowledge access but also the construction of reasoning trajectories, dynamic strategy adjustment, and validation of final results~\citep{zhangcofft}. For effective scientific reasoning, models need not only to reason with available knowledge but also to incorporate key memory patterns. As shown in Figure~\ref{fig_example}, an example from OlympiadBench~\citep{he2024olympiadbench}, GPT-5~\citep{leon2025gpt} without memory fails to consider the kinetic energy theorem, leading to an incorrect reasoning result. However, when memory mechanisms such as the kinetic energy theorem, belt acceleration, and power scenarios are incorporated, the reasoning becomes correct. This illustrates the critical role of memory in enhancing reasoning accuracy and reliability.

\begin{figure*}[t]
	\large
	\centering
	\includegraphics[width=\textwidth]{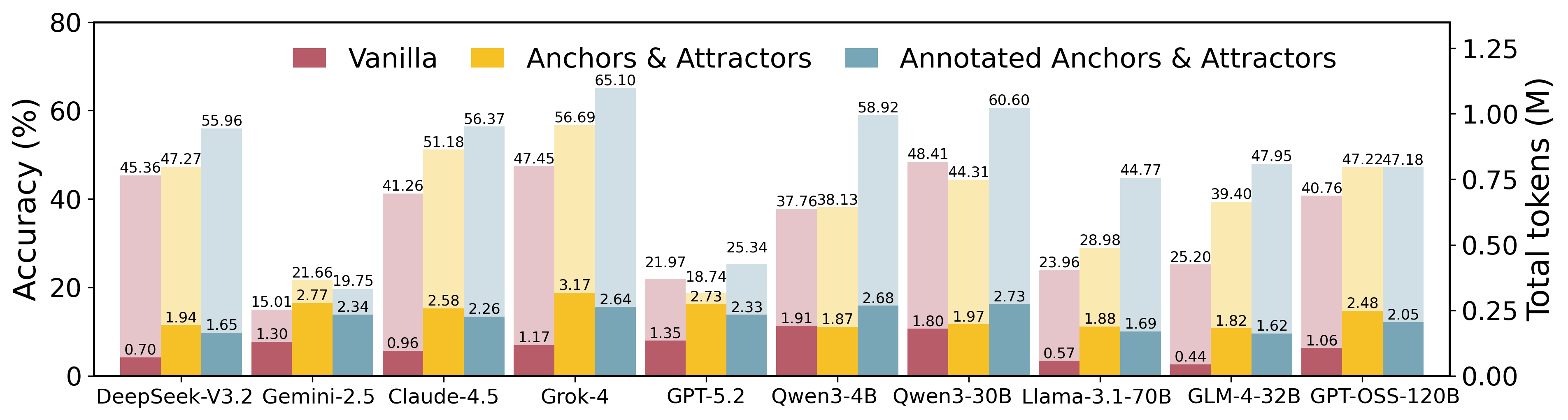}
	\caption{Performance and token analysis across ten LLMs and three memory paradigms. The three color-coded groups represent the experimental paradigms: vanilla, anchors \& attractors, and annotated anchors \& attractors.}
	\label{compare}
\end{figure*}

Existing memory-driven works~\citep{wang2024memoryllm} primarily offer advantages such as memory storage and fast retrieval, efficient knowledge reuse, and reasoning consistency~\citep{xie2024calibrating,cui2024divide} and stability~\citep{liu2025your,du2025investigating}. However, current scientific reasoning benchmarks primarily emphasize final-answer correctness and process consistency, without directly evaluating \textit{memory activation ability}. As a result, they do not reveal whether failures arise from flawed logical inference or from inadequate retrieval and activation of the necessary memory during reasoning.

Human scientific reasoning is closely tied to how memory is organized and accessed. Memory is hierarchically structured, ranging from concrete experiences to abstract schemas~\citep{bein2025schemas}, and relevant knowledge can be selectively activated by contextual cues during problem solving~\citep{liu2012optogenetic}. These properties motivate the construction of benchmarks that align with human memory mechanisms. Such datasets should explicitly represent reusable knowledge units and structured, experience-based templates, and should require context-dependent activation across multiple reasoning steps. Such a benchmark enables fine-grained evaluation of whether models precisely activate the appropriate knowledge and templates at the appropriate time during reasoning, and offers actionable signals to guide the development of more reliable, memory-driven large language models.

To this end, we introduce \textbf{$A^3$-Bench}, a benchmark grounded in \underline{A}nchor and \underline{A}ttractor \underline{A}ctivation and designed to evaluate memory-driven scientific reasoning. Specifically, first, inspired by hierarchical human memory~\citep{bein2025schemas}, we model scientific reasoning memory at two scales: \textit{anchors} (foundational knowledge units) and \textit{attractors} (experience-based templates). Using the SAPM process (\underline{s}ubject, \underline{a}nchor \& attractor, \underline{p}roblem, and \underline{m}emory developing), we annotate 2,198 problems across domains and map each question to its anchor--attractor set. Second, motivated by context-dependent activation in human episodic memory~\citep{liu2012optogenetic}, we introduce a dual-scale memory evaluation framework that leverages anchors and attractors, and propose the \textit{AAUI} (\underline{A}nchor--\underline{A}ttractor \underline{U}tilization \underline{I}ndex) metric to quantify memory activation rates. Third, we conduct experiments on \textbf{$A^3$-Bench} across base models and paradigms, validating its ability to evaluate anchor--attractor memory activation and utilization during reasoning. As shown in Figure~\ref{compare}, this enhancement improves accuracy while keeping token costs controllable, supporting cognitively aligned evaluation and model development.
Our main contributions are as follows:

$\bullet$ We create a dataset of 2,198 scientific reasoning problems via the SAPM process. Each instance is annotated with dual-scale memory signals, \textit{anchors} and \textit{attractors}, reflecting hierarchical human memory across domains.

$\bullet$ $A^3$-Bench is proposed as the first benchmark for memory-driven scientific reasoning. We further propose the \textit{AAUI} metric, which quantifies memory activation rates by leveraging human-like context-dependent episodic recall.

$\bullet$ Experiments validate $A^3$-Bench and show how memory activation shapes multi-step reasoning. This enables fine-grained evaluation of precise memory activation in LLM inference and supports reliable model development.

\section{Preliminaries} \label{preli}

This section introduces the foundational concepts of memory-driven scientific reasoning: \textit{anchor} and \textit{attractor}, \textit{memory activation}, and \textit{memory-augmented reasoning}, which form the theoretical basis for how memory structures guide reasoning.

\paragraph{Definition 1: Anchor and Attractor.}
In reasoning, the \textit{anchor} constrains the initial state and focus the system on relevant knowledge, while the \textit{attractor} represents knowledge structures that guide reasoning along specific paths. Together, the activated anchors and attractors~\citep{zhou2025neural,siegenthaler2025visual} form the state space of the Attractor Basin, unifying these two memory types and describing their collaborative role in reasoning.

\begin{figure*}[t]
	\large
	\centering
	\includegraphics[scale=0.56]{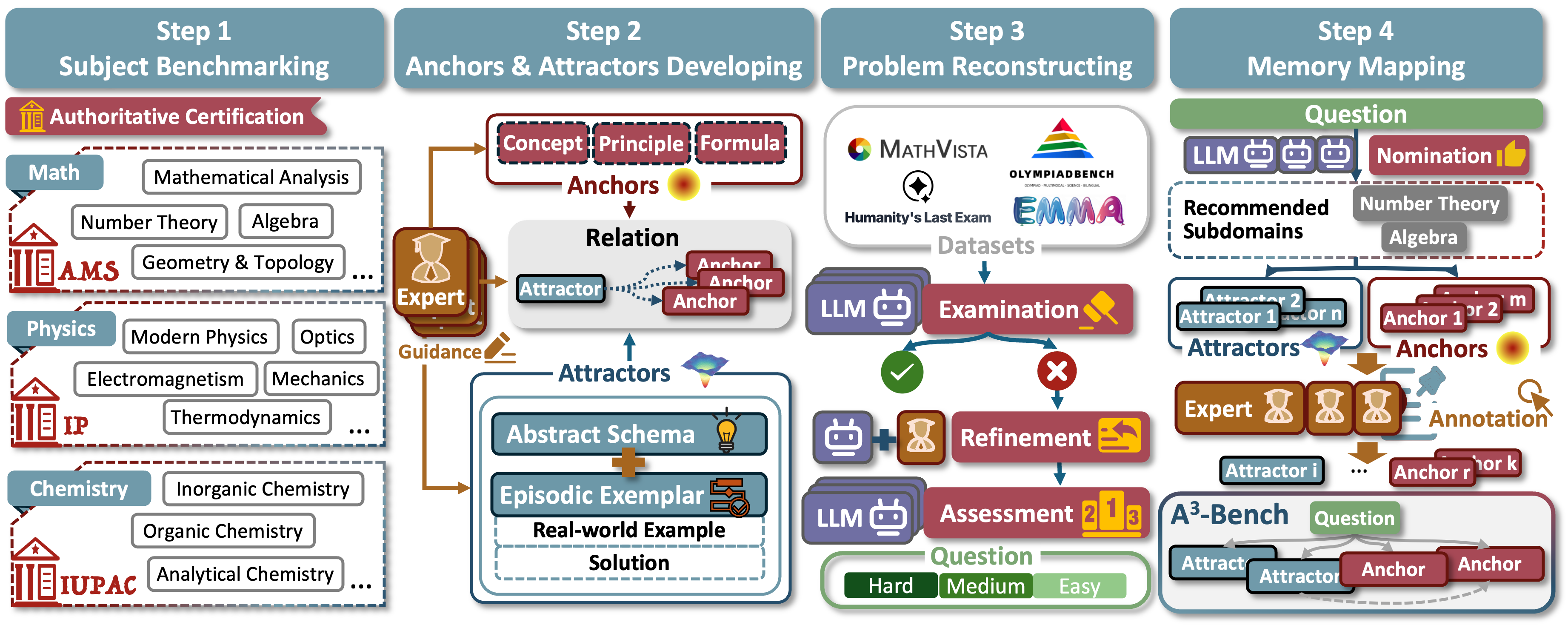}
	\caption{
The four-step annotation process SAPM. First, subject benchmarking defines subdomains for each discipline. Next, experts develop anchors and attractors for each subdomain and define their relations. Then, a new set of questions is refined from existing datasets. Finally, memory mapping associates questions with relevant anchors and attractors.}
	\label{fig_model}
\end{figure*}

Let $\mathcal{Z} \subseteq \mathbb{R}^d$ be a neural or semantic state space, and let $f: \mathcal{Z} \to \mathcal{Z}$ be the dynamical update operator. A state $\mathbf{z}^*$ is an attractor if:
\begin{equation}
\lim_{t \to \infty} f^{(t)}(\mathbf{z}_0) = \mathbf{z}^*.
\end{equation}
The basin of attraction associated with $\mathbf{z}^*$ is defined as:
\begin{equation}
\mathcal{B}(\mathbf{z}^*) = \left\{ \mathbf{z}_0 \in \mathcal{Z} \mid \lim_{t \to \infty} f^{(t)}(\mathbf{z}_0) = \mathbf{z}^* \right\}.
\end{equation}

\paragraph{Definition 2: Memory Activation.}
Memory activation~\citep{friston2010free} is modeled as minimizing the formula:
\begin{equation}
F(\mathbf{z};\mathbf{x})=-\log p(\mathbf{x}\mid\mathbf{z})+D_{\mathrm{KL}}\!\big(q(\mathbf{z})\|p(\mathbf{z})\big),
\end{equation}
where $\mathbf{x}$ denotes the input query, $q(\mathbf{z})$ the posterior representation, and $p(\mathbf{z})$ the prior knowledge distribution. The internal state evolves according to gradient descent:
\begin{equation}
\mathbf{z}_{t+1} = \mathbf{z}_t - \eta \nabla_{\mathbf{z}} F(\mathbf{z}_t; \mathbf{x}),
\end{equation}
driving the system toward an attractor $\mathbf{z}^*$ that best explains the input, corresponding to the activation of memory structures.

\paragraph{Definition 3: Memory-Augmented Reasoning.}
In memory-augmented reasoning~\citep{ko2024memreasoner}, given an input query $s_0$, we map it to an initial internal state $\mathbf{z}_0 = \phi(s_0)$ and identify a set of candidate attractors:
$
\mathcal{A} = \{\mathbf{z}_k^*\}_{k=1}^K
$.
Memory activation is formalized as a minimization problem:
\begin{equation}
\mathbf{z}^* = \arg\min_{\mathbf{z}_k^* \in \mathcal{A}} F(\mathbf{z}_k^*; \mathbf{z}_0).   \label{free-energy}
\end{equation}
Reasoning steps are then guided by the evolving internal representation:
\begin{equation}
s_i \sim \pi_\theta(\cdot \mid s_0, s_{\le i-1}, \mathbf{z}_t),        \label{free-energy-step}
\end{equation}
and the final output is expressed as:
$s_n = \Psi(\mathbf{z}^*, s_0)$,
representing the stable inference outcome after the system settles within the attractor basin associated with the activated knowledge structure.
The proof of memory activation and memory-augmented reasoning is shown in App.~\ref{freeEnergy}.

\section{The \texorpdfstring{$A^{3}$-Bench}{A3-Bench} Dataset} \label{section3}

\begin{figure}[t]
	\large
	\centering
	\includegraphics[width=\columnwidth]{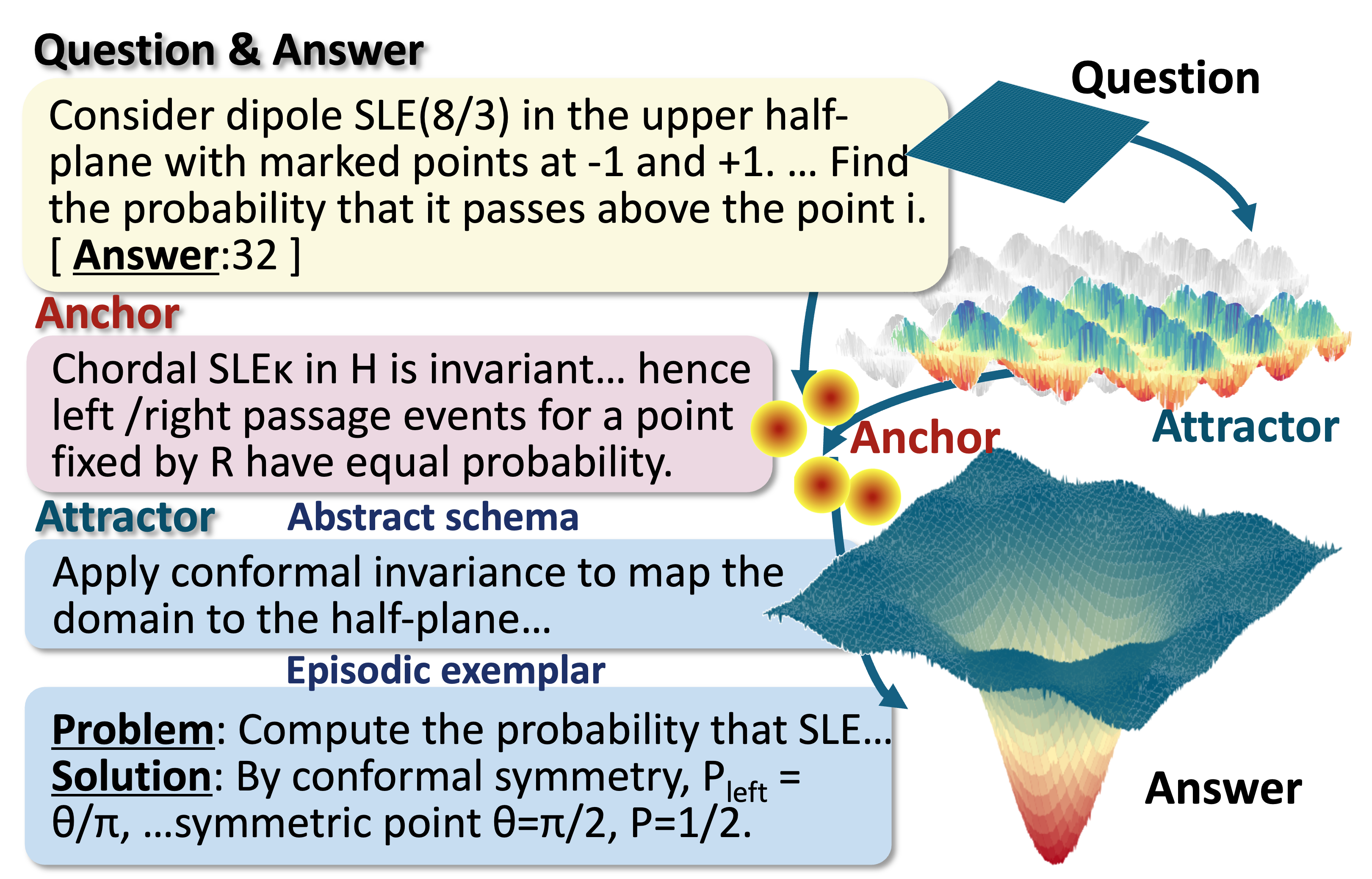}
	\caption{A piece of math problem in $A^3$-Bench.}
	\label{schema}
\end{figure}

This section introduces SAPM, a four-step annotation process for the $A^3$-Bench dataset (Figure~\ref{fig_model}). \S~\ref{3.1} describes subject benchmarking and hierarchical subject standards; \S~\ref{3.2} develops Anchors and Attractors; \S~\ref{3.3} reconstructs problems from existing datasets; and \S~\ref{3.4} performs Memory Mapping by linking questions to anchor--attractor sets. An example data point is shown in Figure~\ref{schema}, and detailed guidance is provided in App.~\ref{annotationGuidance}.


\subsection{Subject Benchmarking} \label{3.1}

Scientific reasoning spans math, physics, and chemistry. For each discipline, we reference authoritative classification systems: math follows the American Mathematical Society (AMS)~\cite{msc2020_report}, physics adopts international standards from the physics community (IP)~\citep{smith2020physics}, and chemistry follows the International Union of Pure and Applied Chemistry (IUPAC)~\citep{heller2013inchi}. We then fine-tune and integrate these systems, resulting in 8 subdomains for math, 5 for physics, and 5 for chemistry. Details are provided in App.~\ref{datasetTaxonomy}.

\subsection{Anchors \& Attractors Developing } \label{3.2}


For each subdomain, we invited three subject experts to label anchors and attractors based on the established subdomains, following our memory development guidelines. Anchors include concepts, principles, and formulas, which set initial conditions and guide reasoning. Attractors, including abstract schemas and specific exemplars, ensure reasoning unfolds within a predefined framework.

\subsection{Problem Reconstructing} \label{3.3}


In this section, we construct a new problem set from existing datasets in the following stages:

\paragraph{Examination} Given the varying difficulty levels and task focus of the four datasets MathVista~\citep{lu2023mathvista}, OlympiadBench, EMMA~\citep{hao2025can}, and Humanity's Last Exam~\citep{phan2025humanity}, we begin by examining the questions \( Q = \{q_1, q_2, \dots, q_n\} \). Each question \( q_i \) is answered by three LLMs: GPT-5, Deepseek-V3.2~\citep{liu2024deepseek}, and Qwen-30B~\citep{yang2025qwen3technicalreport}. 
The diversity in model parameters, capabilities, and balance between open-source and closed-source models ensures varied responses. A question is passed to the next phase if any model answers incorrectly, and discarded if all models answer correctly.

\begin{figure}[t]
	\large
	\centering
	\includegraphics[width=\columnwidth]{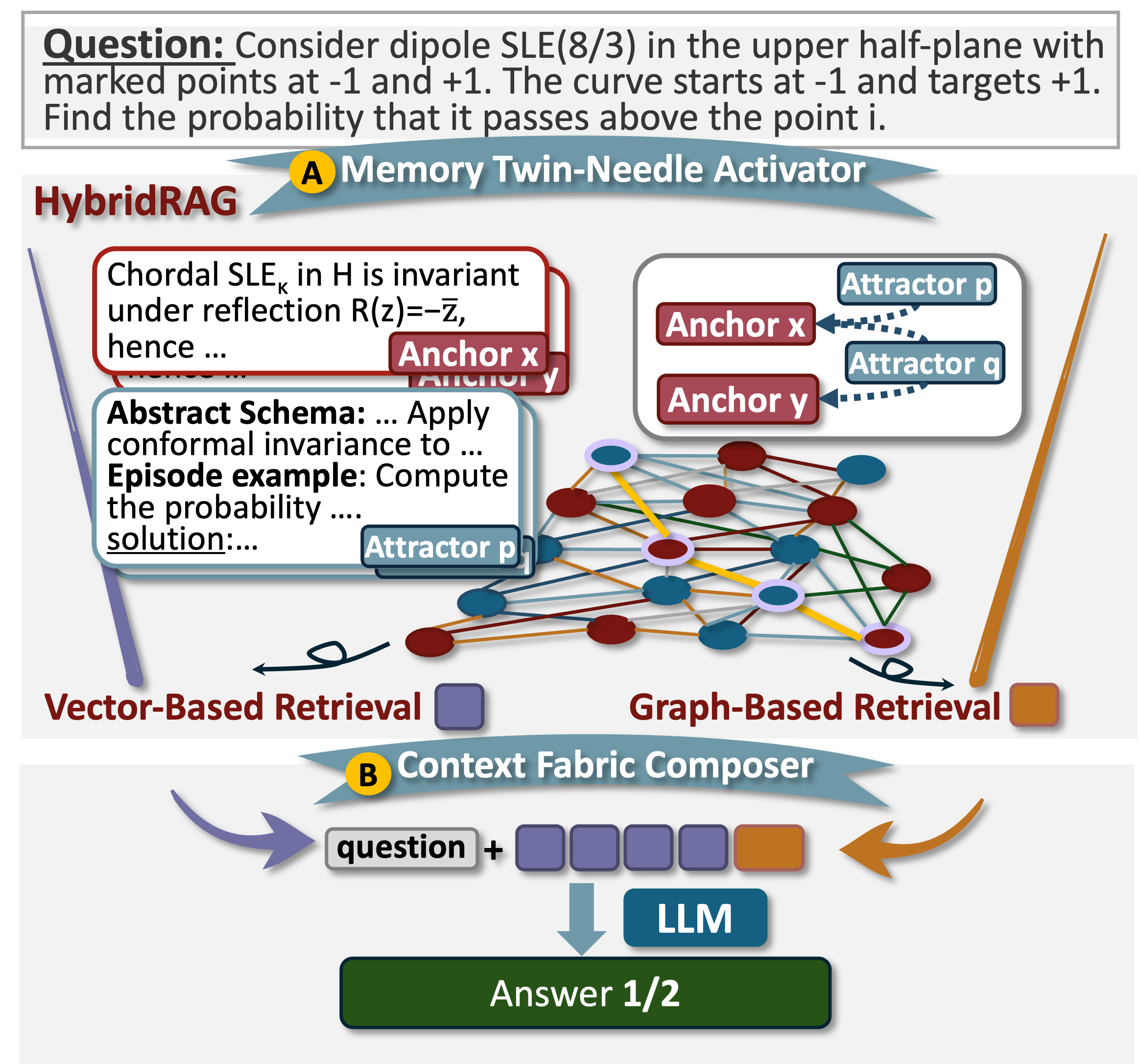}
	\caption{Schema of the $A^3$-Bench dataset and its usage within a HybridRAG framework. (a) memory twin-needle activator. (b) context fabric composer.}
	\label{hybridrag}
\end{figure}

\begin{table*}[ht]
\centering
\setlength{\tabcolsep}{3pt}
\resizebox{\textwidth}{!}{
\begin{tabular}{l|cccc|cccc|cccc|ccc}
\toprule
\multirow{2}{*}{\textbf{Method}} &
\multicolumn{4}{c|}{\textbf{Math}} &
\multicolumn{4}{c|}{\textbf{Physics}} &
\multicolumn{4}{c|}{\textbf{Chemistry}} &
\multirow{2}{*}{\textbf{Avg.}} &
\multirow{2}{*}{\textbf{AAUI}} &
\multirow{2}{*}{\textbf{Tokens}} \\
& \textbf{Easy} & \textbf{Medium} & \textbf{Hard} & \textbf{Avg.}
& \textbf{Easy} & \textbf{Medium} & \textbf{Hard} & \textbf{Avg.}
& \textbf{Easy} & \textbf{Medium} & \textbf{Hard} & \textbf{Avg.}
& & & \\
\midrule

\rowcolor{lightgray}
\multicolumn{16}{c}{\textbf{Vanilla}} \\
\midrule
DeepSeek-V3.2 & 46.37 & 39.46 & 26.33 & 38.28 & \textbf{64.58} & \underline{53.33} & \textbf{31.67} & \underline{51.33} & 60.42 & 53.33 & 36.67 & 51.17 & 45.36 & 0 & $7.04\times10^{5}$ \\
Gemini-2.5-Flash & 24.81 & 31.44 & 22.67 & 26.15 & 3.75 & 5.00 & 5.00 & 4.50 & 5.00 & 3.89 & 12.78 & 7.00 & 15.01 & 0 & $1.30\times10^{6}$ \\
Claude-Haiku-4.5 & 43.11 & 38.46 & 27.67 & 37.07 & 50.42 & 36.11 & 21.11 & 37.33 & \textbf{64.58} & \underline{54.44} & 33.33 & \underline{52.17} & 41.26 & 0 & $9.57\times10^{5}$ \\
Grok-4-Fast & \underline{51.38} & \underline{44.82} & 33.33 & \underline{43.99} & \underline{62.50} & \textbf{58.33} & \underline{30.00} & \textbf{51.50} & 58.75 & 48.33 & 37.22 & 49.17 & \underline{47.45} & 0 & $1.18\times10^{6}$ \\
GPT-5-Mini & 31.83 & 25.42 & 21.00 & 26.65 & 12.92 & 11.11 & 12.22 & 12.17 & 26.67 & 26.11 & 18.33 & 24.00 & 21.97 & 0 & $1.35\times10^{6}$ \\
Qwen3-4B & 47.62 & 44.48 & \underline{35.00} & 42.89 & 29.58 & 20.56 & 17.78 & 23.33 & 46.67 & 44.44 & \underline{38.89} & 43.67 & 37.76 & 0 & $1.91\times10^{6}$ \\
Qwen3-30B & \textbf{55.64} & \textbf{52.17} & \textbf{36.67} & \textbf{48.90} & 56.25 & 41.11 & 28.33 & 43.33 & \underline{60.83} & \textbf{56.11} & 38.33 & \textbf{52.67} & \textbf{48.41} & 0 & $1.81\times10^{6}$ \\
Llama-3.1-70B & 34.21 & 27.40 & 17.53 & 26.94 & 27.03 & 16.04 & 11.81 & 18.90 & 28.21 & 24.39 & 26.67 & 26.83 & 23.96 & 0 & $5.74\times10^{5}$ \\
GLM-4-32B & 37.59 & 26.76 & 18.33 & 28.56 & 27.08 & 21.67 & 14.44 & 21.67 & 30.00 & 23.33 & 13.89 & 23.17 & 25.20 & 0 & $4.40\times10^{5}$ \\
GPT-OSS-120B & 49.12 & 38.13 & 30.33 & 40.18 & 47.50 & 30.00 & 26.67 & 36.00 & 50.42 & 43.89 & \textbf{43.89} & 46.50 & 40.76 & 0 & $4.40\times10^{5}$ \\
\midrule

\rowcolor{lightgray}
\multicolumn{16}{c}{\textbf{+ Anchor \& Attractor Activation}} \\
\midrule
DeepSeek-V3.2 & 59.40 & 54.52 & 30.00 & 49.10 & \underline{62.92} & \underline{58.89} & \underline{35.00} & \underline{53.33} & 44.17 & 38.89 & 29.44 & 38.17 & 47.27 & 0.22 & $1.94\times10^{6}$ \\
Gemini-2.5-Flash & 30.58 & 36.45 & 23.67 & 30.26 & 10.00 & 3.89 & 6.67 & 7.17 & 23.75 & 18.33 & 22.78 & 21.83 & 21.66 & 0.14 & $2.77\times10^{6}$ \\
Claude-Haiku-4.5 & \underline{64.66} & \textbf{56.86} & 33.00 & \underline{52.81} & 60.42 & 56.11 & 31.11 & 50.33 & \underline{62.50} & \underline{47.22} & \underline{33.89} & \underline{49.33} & \underline{51.18} & 0.46 & $2.58\times10^{6}$ \\
Grok-4-Fast & \textbf{68.92} & \textbf{57.19} & \textbf{39.33} & \textbf{56.51} & \textbf{72.50} & \textbf{63.89} & \textbf{41.67} & \textbf{60.67} & \textbf{65.00} & \textbf{56.11} & \textbf{33.89} & \textbf{53.00} & \textbf{56.69} & 0.66 & $3.17\times10^{6}$ \\
GPT-5-Mini & 26.32 & 25.08 & 21.00 & 24.35 & 11.67 & 12.22 & 11.67 & 11.83 & 18.33 & 18.33 & 11.67 & 16.33 & 18.74 & 0.09 & $2.74\times10^{6}$ \\
Qwen3-4B & 59.15 & 46.15 & 30.33 & 46.59 & 47.50 & 43.89 & 23.33 & 39.17 & 30.83 & 20.00 & 15.56 & 23.00 & 38.13 & 0.27 & $1.87\times10^{6}$ \\
Qwen3-30B & 64.16 & 51.84 & 33.33 & 51.20 & 60.42 & 46.67 & 27.78 & 46.50 & 39.17 & 28.33 & 21.67 & 30.67 & 44.31 & 0.36 & $1.97\times10^{6}$ \\
Llama-3.1-70B & 44.11 & 39.13 & 27.00 & 37.47 & 33.33 & 24.44 & 13.89 & 24.83 & 20.83 & 18.33 & 17.33 & 19.00 & 28.98 & 0.33 & $1.88\times10^{6}$ \\
GLM-4-32B & 59.90 & 50.84 & 30.33 & 48.30 & 52.50 & 43.33 & 18.33 & 39.50 & 29.58 & 21.11 & 21.11 & 24.50 & 39.40 & 0.41 & $1.82\times10^{6}$ \\
GPT-OSS-120B & 56.14 & 50.17 & \underline{36.67} & 48.50 & 57.08 & 46.11 & 35.00 & 47.17 & 52.08 & 43.33 & 37.78 & 45.17 & 47.22 & 0.44 & $2.48\times10^{6}$ \\
\midrule

\rowcolor{lightgray}
\multicolumn{16}{c}{\textbf{+ Annotated Anchor \& Attractor Activation}} \\
\midrule
DeepSeek-V3.2 & 65.66 & 57.19 & 32.33 & 53.11 & \underline{72.50} & 60.00 & 35.00 & 57.50 & 73.75 & 59.44 & 39.44 & 59.17 & 55.96 & 0.88 & $1.65\times10^{6}$ \\
Gemini-2.5-Flash & 37.84 & 39.46 & 27.00 & 35.07 & 1.25 & 3.89 & 5.56 & 3.33 & 8.33 & 8.89 & 15.56 & 10.67 & 19.75 & 0.69 & $2.34\times10^{6}$ \\
Claude-Haiku-4.5 & 70.93 & 58.53 & 34.33 & 56.21 & 63.75 & 52.78 & 34.44 & 51.67 & \underline{74.58} & 62.22 & 42.78 & 61.33 & 56.37 & 0.77 & $2.26\times10^{6}$ \\
Grok-4-Fast & \textbf{75.94} & 59.20 & 40.00 & 60.12 & \textbf{80.00} & \textbf{79.44} & \textbf{55.00} & \textbf{72.33} & \textbf{78.75} & \textbf{70.56} & \underline{45.00} & \textbf{66.17} & \textbf{65.10} & 0.97 & $2.64\times10^{6}$ \\
GPT-5-Mini & 36.59 & 28.09 & 31.67 & 32.57 & 16.67 & 16.11 & 15.00 & 16.00 & 22.92 & 22.78 & 22.22 & 22.67 & 25.34 & 0.74 & $2.33\times10^{6}$ \\
Qwen3-4B & 72.18 & \textbf{63.21} & \underline{45.67} & \underline{61.52} & 60.83 & 53.33 & 34.44 & 50.67 & 69.17 & \underline{66.11} & \textbf{51.11} & \underline{62.83} & 58.92 & 0.92 & $2.68\times10^{6}$ \\
Qwen3-30B & \underline{73.18} & \underline{62.88} & \textbf{47.33} & \textbf{62.32} & 67.08 & \underline{62.22} & \underline{42.78} & \underline{58.33} & 71.67 & 63.33 & 41.11 & 60.00 & \underline{60.60} & 0.95 & $2.73\times10^{6}$ \\
Llama-3.1-70B & 56.64 & 48.83 & 33.00 & 47.19 & 45.83 & 41.11 & 21.11 & 37.00 & 55.00 & 51.11 & 37.22 & 48.50 & 44.77 & 0.96 & $1.69\times10^{6}$ \\
GLM-4-32B & 63.91 & 53.18 & 34.67 & 51.90 & 55.83 & 45.56 & 30.00 & 45.00 & 55.42 & 43.33 & 30.56 & 44.33 & 47.95 & 0.92 & $1.62\times10^{6}$ \\
GPT-OSS-120B & 59.40 & 46.49 & 34.00 & 47.90 & 47.08 & 38.89 & 28.33 & 39.00 & 63.33 & 53.89 & 42.22 & 54.17 & 47.18 & 0.68 & $2.05\times10^{6}$ \\
\bottomrule
\end{tabular}}
\caption{Main results on $A^{3}$-Bench under different memory paradigms across ten LLMs.}
\label{tab:main_results}
\end{table*}

\paragraph{Refinement}

For questions moving to the second stage, three LLMs \( \mathcal{M} \) perform a cross-analysis. Each model \( \mathcal{M}_j \) solves the problem, while the other two models \( \mathcal{M}_k \) and \( \mathcal{M}_l \) evaluate its reasoning and identify errors. Let \( A_j(q_i) \) be the answer from model \( \mathcal{M}_j \) for question \( q_i \). The evaluation function \( E_{k,l}\) represents the errors detected:
\begin{equation}
E_{k,l}(A_j, q_i) = \sum_{e=1}^m \text{Error}_e(A_j, q_i),    
\end{equation}
where \( \text{Error}_e(A_j, q_i) \) returns indictor if error \( e \) is found. After this, three subject experts revise the question \( q_i \) to \( q_i' \), integrating multi-step reasoning. The revision function is \( q_i' = f(q_i, E_{k,l},R) \), where \( R \) represents the reasoning steps. A standard answer \( A'(q_i') \) is provided for each revised question.

\paragraph{Assessment}

Once revisions are complete, the new problems are evaluated by the three LLMs. The same three models \( \mathcal{M} \) answer each question \( q_i \) 10 times, resulting in 30 answers per question. The difficulty is based on the number of correct answers. Let \( C = \text{Correct}(A_j, q_i) \) return 1 for a correct answer and 0 for an incorrect one. A question is classified as "Easy" if \( 15 \leq \sum_{j} C \leq 30 \), "Medium" if \( 5 \leq \sum_{j} C \leq 14 \), and "Difficult" if \( 0 \leq \sum_{j} C \leq 4 \). The overall difficulty is determined by the majority of correct answers.

\subsection{Memory Mapping} \label{3.4}

Based on the question pool and anchor--attractor library, the process proceeds as follows:

First, three LLMs \( \mathcal{M} \) recommend a subdomain for each question \( q_i \). If at least two models agree, the subdomain is finalized through a voting mechanism: \( \text{Vote} \left( \mathcal{M}_1(q_i), \mathcal{M}_2(q_i), \mathcal{M}_3(q_i) \right) \).

Next, human experts review the recommendations and manually annotate relevant anchors and attractors, ensuring they are strongly related to the reasoning process and belong to the same subdomain. For each question, the anchors and attractors are denoted as \( \{\text{Anchor}_i\}_{i=1}^{n_a} \) and \( \{\text{Attractor}_i\}_{i=1}^{n_t} \), where \( n_a \le 6 \) and \( n_t \le 4 \).

Finally, the \( A^3 \)-Bench problem set is created and associated with the anchors and attractors. The statistics of \( A^3 \)-Bench are shown in Table~\ref{statistics}.

\begin{table}[ht]
\centering
\small
\begin{tabular}{l r}
\toprule
\textbf{Statistics} & \textbf{Number} \\ 
\midrule
Total Problems & 2,198 \\ 
\midrule
\multicolumn{2}{l}{\textbf{By Subject}} \\
Math & 998(45.40\%) \\
Physics & 600(27.30\%) \\
Chemistry & 600(27.30\%) \\
\midrule
\multicolumn{2}{l}{\textbf{By Difficulty}} \\
Easy & 879(40.00\%) \\
Medium & 659(29.98\%) \\
Hard & 660(30.02\%) \\
\midrule
\multicolumn{2}{l}{\textbf{Anchors/Attractors (per problem)}} \\
Average Anchor Count & 2.79 \\
Average Attractor Count & 2.33 \\
Max Anchor Count & 6 \\
Max Attractor Count & 4 \\
\bottomrule
\end{tabular}
\caption{Statistics of $A^3$-Bench.}
\label{statistics}
\end{table}

\begin{table}[ht]
\centering
\Large
\resizebox{\columnwidth}{!}{
\begin{tabular}{l|cccc|cc}
\toprule
\textbf{Method} &
\textbf{M.C.} &
\textbf{M.E.} &
\textbf{P.C.} &
\textbf{P.E.} &
\textbf{Avg.} &
\textbf{Tokens} \\
\midrule
\rowcolor{lightgray}
\multicolumn{7}{c}{\textbf{Vanilla}} \\
\midrule
DeepSeek-V3.2 & 15.34 & \underline{27.02} & 11.02 & \underline{35.65} & 21.25 & $1.12\times10^{6}$ \\
Gemini-2.5-Flash & 6.47 & 20.32 & 1.27 & 26.09 & 13.28 & $1.74\times10^{6}$ \\
Claude-Haiku-4.5 & \textbf{32.44} & \textbf{49.27} & \underline{21.61} & \textbf{55.65} & \textbf{40.29} & $1.58\times10^{6}$ \\
Grok-4-Fast & \underline{27.54} & 26.61 & \textbf{28.39} & \underline{35.65} & \underline{27.53} & $2.41\times10^{6}$ \\
GPT-5-Mini & 18.11 & 13.95 & 5.93 & 4.35 & 14.52 & $1.63\times10^{6}$ \\
Qwen3-4B & 7.02 & 13.39 & 1.69 & 14.78 & 9.84 & $1.14\times10^{6}$ \\
Qwen3-30B & 9.15 & 20.40 & 4.66 & 25.22 & 14.67 & $1.03\times10^{6}$ \\
Llama-3.1-70B & 3.33 & 6.85 & 2.97 & 10.43 & 5.24 & $1.32\times10^{6}$ \\
GLM-4-32B & 8.69 & 15.32 & 3.39 & 18.26 & 11.71 & $9.85\times10^{5}$ \\
GPT-OSS-120B & 11.09 & 22.10 & 8.90 & 29.57 & 16.80 & $1.61\times10^{6}$ \\
\midrule
\rowcolor{lightgray}
\multicolumn{7}{c}{\textbf{Chain of Thought}} \\
\midrule
DeepSeek-V3.2 & 20.61 & \underline{33.47} & 14.83 & \underline{41.74} & 26.97 & $2.28\times10^{6}$ \\
Gemini-2.5-Flash & 10.91 & 24.27 & 5.08 & 31.30 & 17.47 & $3.61\times10^{6}$ \\
Claude-Haiku-4.5 & \textbf{38.63} & \textbf{55.08} & \underline{26.69} & \textbf{60.87} & \textbf{46.17} & $3.83\times10^{6}$ \\
Grok-4-Fast & \underline{33.92} & 31.13 & \textbf{33.47} & 40.87 & \underline{32.88} & $5.21\times10^{6}$ \\
GPT-5-Mini & 22.92 & 17.82 & 10.17 & 6.09 & 18.71 & $3.72\times10^{6}$ \\
Qwen3-4B & 10.53 & 18.47 & 3.81 & 20.00 & 14.03 & $2.46\times10^{6}$ \\
Qwen3-30B & 12.94 & 26.37 & 8.90 & 31.30 & 19.60 & $2.18\times10^{6}$ \\
Llama-3.1-70B & 5.91 & 10.40 & 5.93 & 14.78 & 8.38 & $2.63\times10^{6}$ \\
GLM-4-32B & 12.85 & 20.81 & 5.93 & 25.22 & 16.46 & $2.39\times10^{6}$ \\
GPT-OSS-120B & 16.17 & 27.74 & 13.56 & 35.65 & 22.15 & $3.47\times10^{6}$ \\
\midrule
\rowcolor{lightgray}
\multicolumn{7}{c}{\textbf{Anchor \& Attractor Activation}} \\
\midrule
DeepSeek-V3.2 & 27.08 & \underline{40.24} & 21.19 & \underline{48.70} & 33.60 & $2.35\times10^{6}$ \\
Gemini-2.5-Flash & 17.74 & 29.11 & 10.59 & 38.26 & 23.27 & $3.53\times10^{6}$ \\
Claude-Haiku-4.5 & \textbf{45.38} & \textbf{61.69} & \underline{32.63} & \textbf{67.83} & \textbf{52.79} & $3.57\times10^{6}$ \\
Grok-4-Fast & \underline{40.66} & 39.11 & \textbf{39.41} & 47.83 & \underline{40.14} & $4.98\times10^{6}$ \\
GPT-5-Mini & 29.39 & 23.87 & 16.53 & 12.17 & 24.95 & $3.46\times10^{6}$ \\
Qwen3-4B & 16.91 & 24.76 & 7.63 & 26.96 & 20.16 & $2.31\times10^{6}$ \\
Qwen3-30B & 19.13 & 32.66 & 15.25 & 38.26 & 25.89 & $2.33\times10^{6}$ \\
Llama-3.1-70B & 11.46 & 16.05 & 11.44 & 20.87 & 13.99 & $2.37\times10^{6}$ \\
GLM-4-32B & 19.22 & 27.50 & 11.86 & 32.17 & 22.97 & $2.25\times10^{6}$ \\
GPT-OSS-120B & 22.73 & 33.95 & 20.34 & 42.61 & 28.58 & $3.15\times10^{6}$ \\
\bottomrule
\end{tabular}}
\caption{Generalized experiments for OlympiadBench.}
\label{gen_results}
\end{table}

\section{Experiments}
This section outlines the experimental framework for focusing on the accurate activation of memory underlying scientific reasoning. \S~\ref{4.1} describes the benchmarking method, \S~\ref{4.2} introduces memory paradigms, and \S~\ref{4.3} presents the evaluation synergy metric.

\subsection{Benchmarking Method} \label{4.1}
We instantiate the proposed memory-activation method by adapting HybridRAG~\citep{sarmah2024hybridragintegratingknowledgegraphs}. As shown in Figure~\ref{hybridrag}, the framework consists of two core components: the Memory Twin-Needle Activator and the Context Fabric Composer.

\paragraph{Memory Twin-Needle Activator.}
We build a hybrid memory substrate: Anchors/Attractors are indexed in a dense store $\mathcal{I}_{\mathrm{vec}}$ and organized in a knowledge graph $G=(V_{\mathrm{anc}}\cup V_{\mathrm{attr}},E_{\mathrm{rel}})$. 
Instantiated with HybridRAG, the \emph{Vector Needle} retrieves top-$k$ nodes by semantic similarity, while the \emph{Graph Needle} traverses $E_{\mathrm{rel}}$ to recover their logical links:
\begin{equation}
\mathbf{z}^* \approx \Phi_{\mathrm{hybrid}}(x)
\triangleq
\mathcal{V}(x)\,\oplus\,\mathcal{G}\!\big(\mathcal{V}(x)\big).
\end{equation}

\paragraph{Context Fabric Composer.}
We compose the final context by weaving the query $x$ with the activated state $\mathbf{z}^*$:
\begin{equation}
C_{\mathrm{final}}=\mathcal{W}(x,\mathbf{z}^*)
\triangleq
\mathcal{I}\,\oplus\,\Big[x \,\bowtie\, \mathcal{S}(\mathbf{z}^*)\Big],
\end{equation}
where $\mathcal{I}$ is a fixed instruction prefix, $\mathcal{S}(\cdot)$ serializes $\mathbf{z}^*$ into an LLM-readable form.

\subsection{Memory Paradigms} \label{4.2}

\paragraph{Paradigms.}

We evaluate memory-driven scientific reasoning under three paradigms: (i) \emph{No memory}, where the model answers from parametric knowledge $A=\mathcal{M}(Q)$; (ii) \emph{Full memory}, where it conditions on activated evidence from the full library $A=\mathcal{M}\!\big(Q;\,\mathrm{Activate}(Q,\mathcal{K}_{Total})\big)$; and (iii) \emph{Gold memory}, which restricts activation to the human-labeled subset $A=\mathcal{M}\!\big(Q;\,\mathrm{Activate}(Q,\mathcal{K}_{Gold})\big)$.

\paragraph{Base Models.}
We choose 10 LLMs spanning scales, architectures, and access types (open vs.\ proprietary): DeepSeek-V3.2, Gemini-2.5-Flash~\citep{comanici2025gemini25pushingfrontier}, Claude-Haiku-4.5~\citep{anthropic_claude_haiku_4_5_2025}, Grok-4-Fast~\citep{xai_grok_4_fast_2025}, GPT-5-Mini, Qwen3-4B, Qwen3-30B, Llama-3.1-70B~\citep{grattafiori2024llama}, GLM-4-32B~\citep{glm2024chatglm}, and GPT-OSS-120B.

\subsection{Evaluation Metrics} \label{4.3}

\paragraph{Accuracy (Acc).}
We report Acc by matching the model's final answer to the ground truth.

\paragraph{AAUI.}
We propose \textbf{AAUI} (Anchor--Attractor Utilization Index) to measure how well a model activates expert-annotated Anchors/Attractors during reasoning. For annotated Anchors and Attractors sets $A_i,T_i$ and response $y_i$, define
$AU_i = \frac{1}{|A_i|}\sum_{a\in A_i}\mathbb{1}_A(a,y_i)$,
$TU_i = \frac{1}{|T_i|}\sum_{t\in T_i}\mathbb{1}_T(t,y_i)$,
where $\mathbb{1}$ indicates semantic presence in $y_i$. We compute
\begin{equation}
\text{AAUI}_i=\frac{1}{2}\left(\frac{AU_i+TU_i}{2}+AU_i\cdot TU_i\right), 
\end{equation}
where $\text{AAUI}_i \in [0,1]$.
$\text{AAUI}=\frac{1}{N}\sum_{i=1}^{N}\text{AAUI}_i$,
which combines anchor/attractor recall with an interaction term to reward simultaneous activation.

\begin{figure}[t]
	\large
	\centering
	\includegraphics[width=\columnwidth]{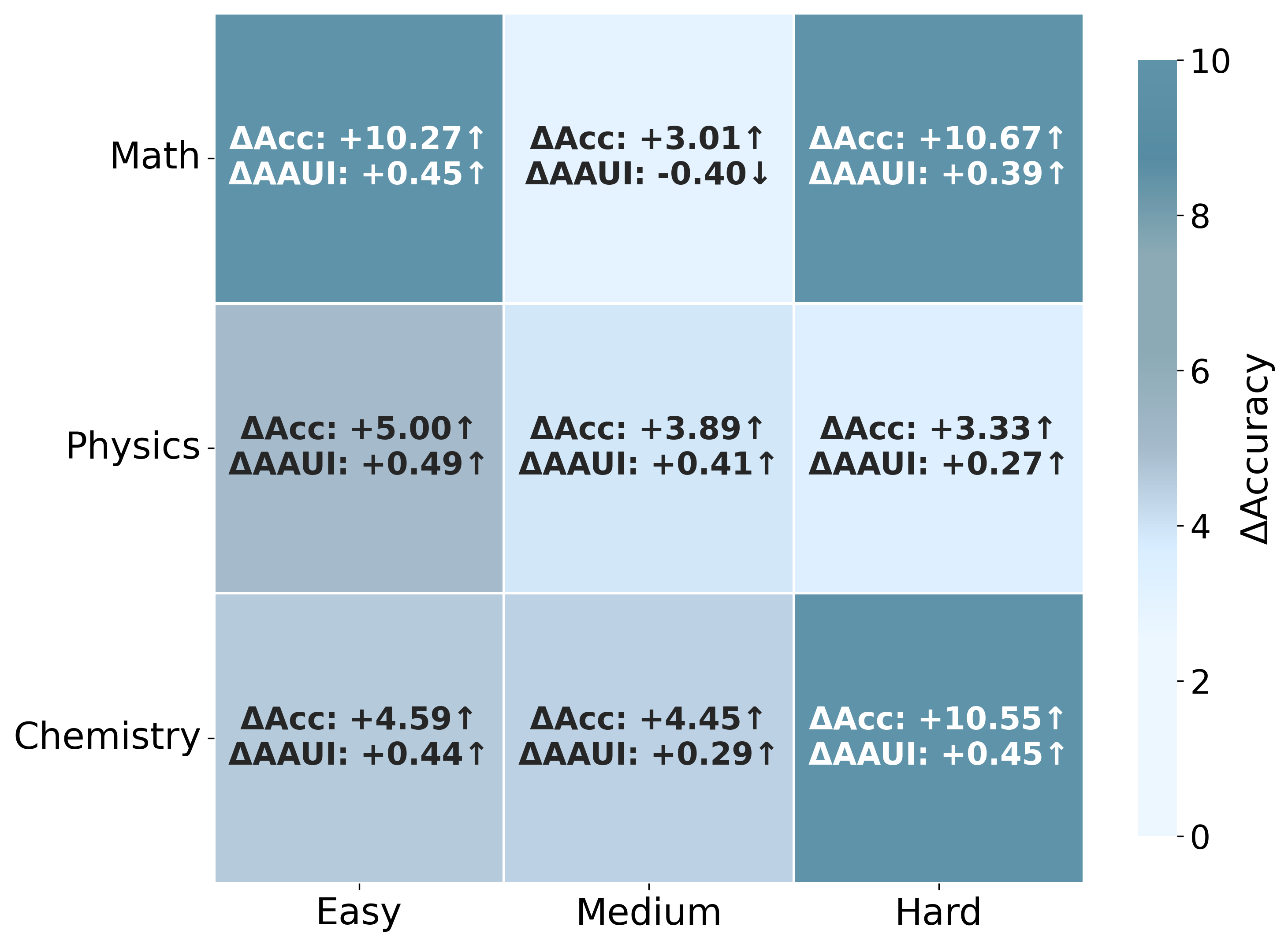}
	\caption{Heatmap analysis of performance gains and memory utility across subjects and difficulties. 
    }
	\label{heatmap}
\end{figure}

\section{Analysis}

This section presents the main results (\S~\ref{5.1}), generalized analysis (\S~\ref{5.2}), memory gains (\S~\ref{5.3}), inference-time analysis (\S~\ref{5.4}), and error-type distribution (\S~\ref{5.5}). App.~\ref{differnetParadigms} evaluates anchor-only and attractor-only activation, App.~\ref{otherAnalysis} reports significance tests and noise interference, and App.~\ref{caseStudy} provides successful and failure cases.

\subsection{Main Results} \label{5.1}

Table~\ref{tab:main_results} reports the performance of ten LLMs under three paradigms. The results provide four main findings about memory-driven scientific reasoning.

\paragraph{Memory augmentation consistently improves scientific reasoning across LLMs and subjects.}
Across subjects and difficulty levels, all ten LLMs achieve higher accuracy under Annotated Activation than in the Vanilla, improving the average from 34.71\% to 48.19\% (+13.48). The gains are model-dependent, ranging from modest increases (e.g., GPT-5-Mini +3.37; Gemini-2.5-Flash +4.74) to substantial boosts (e.g., GLM-4-32B +22.75; Qwen3-4B +21.16; Llama-3.1-70B +20.81), indicating heterogeneous ability to leverage activated Anchors and Attractors.



\paragraph{Memory activation is most beneficial on hard problems and reduces the difficulty gap.}
In the Vanilla paradigm, Hard subsets remain difficult for most models (e.g., Physics-Hard: Grok-4-Fast 30.00\%, GLM-4-32B 14.44\%, Qwen3-4B 17.78\%, Llama-3.1-70B 11.81\%).
With annotated memory, Hard performance improves substantially, especially in Physics (e.g., Grok-4-Fast +25.00, GLM-4-32B +15.56), and also in other domains (e.g., Qwen3-4B: Math-Hard +10.67, Chemistry-Hard +12.22).
These gains suggest that many hard problems fail due to missing or misselected solution templates (Attractors); activating the right template makes multi-step reasoning more tractable.

\paragraph{AAUI correlates with accuracy and diagnoses reasoning fidelity under memory activation.}
Under the Anchor \& Attractor Activation paradigm, higher AAUI generally aligns with higher accuracy.
For example, Grok-4-Fast reaches AAUI$=0.66$ with Avg.$=56.69\%$ and Claude-Haiku-4.5 reaches AAUI$=0.46$ with Avg.$=51.18\%$, while GPT-5-Mini has AAUI$=0.09$ with Avg.$=18.74\%$.
This pattern suggests that AAUI captures whether a model co-activates compatible Anchors and Attractors and converts them into correct reasoning.

\begin{figure}[t]
	\large
	\centering
	\includegraphics[width=\columnwidth]{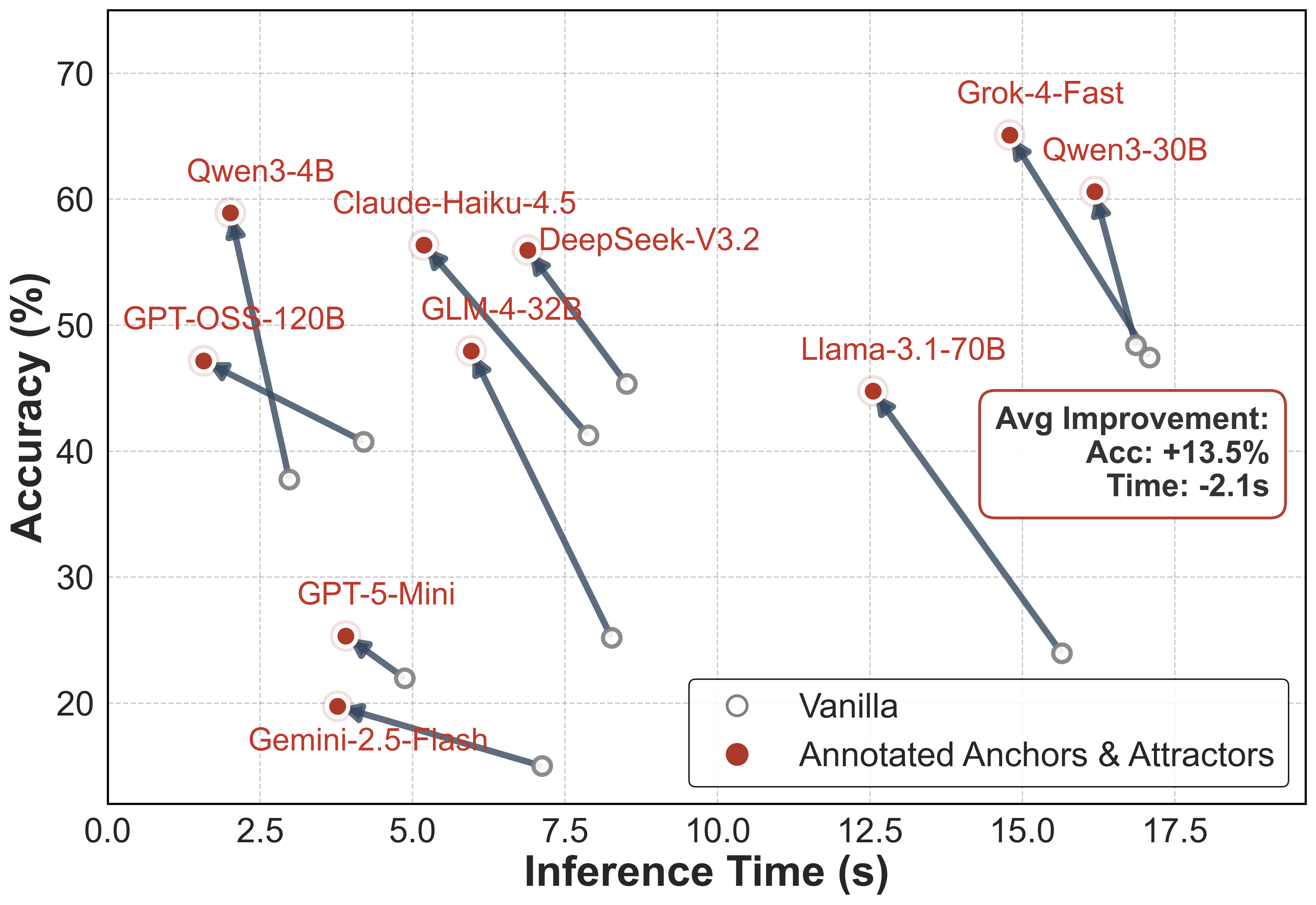}
	\caption{Acc. (\%) vs. avg. inference time (s) per question. Gray: vanilla; Red: annotated anchors \& attractors. Arrows indicate the performance shift.}
	\label{time}
\end{figure}

\subsection{Generalized Analysis} \label{5.2}

To assess the transferability of our memory mechanism, we evaluate the same models on the OlympiadBench, which includes competition-level (M.C., P.C.) and entrance-exam-level (M.E., P.E.) problems. Table~\ref{gen_results} summarizes the results and supports two observations.

\paragraph{Anchor--Attractor activation generalizes beyond the source dataset.}
For all ten models, Anchor--Attractor Activation consistently outperforms both the Vanilla baseline and CoT prompting.
Overall, Anchor \& Attractor Activation improves the average score across models by 11.12 points over Vanilla and by 6.35 points over CoT. For example, DeepSeek-V3.2 reaches 33.60\% with activation (+12.35 vs.\ Vanilla; +6.63 vs.\ CoT).
This suggests our Anchors and Attractors capture reusable scientific concepts and solution patterns that transfer to unseen high-difficulty problems.

\paragraph{Gains are largest on competition-level subsets.}
Improvements are most pronounced on the hardest competition subsets.
On Physics Competition (P.C.), where Vanilla performance is extremely low (e.g., Qwen3-4B: 1.69\%), activation raises the score to 7.63\% ($\sim$4.5$\times$).
Similarly, Claude-Haiku-4.5 improves from 21.61\% (Vanilla) to 32.63\% (Activated) on P.C.
These results indicate that CoT may help derivations but often misses the right starting principles; activating Attractors and supporting Anchors helps recover viable solution paths.

\begin{figure}[t]
	\large
	\centering
	\includegraphics[width=\columnwidth]{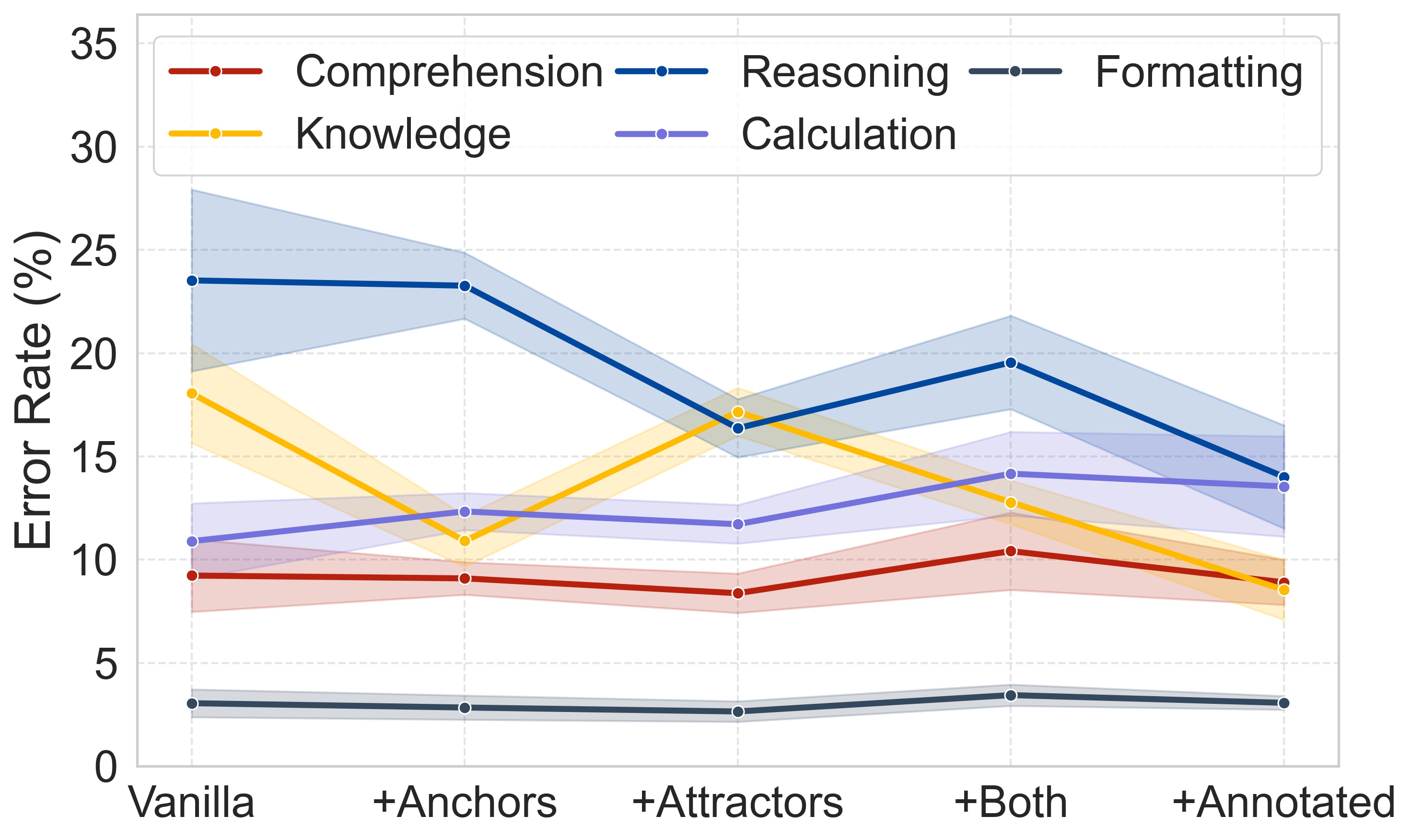}
	\caption{Evolution of error distributions across five paradigms.}
	\label{errorType}
\end{figure}

\subsection{Gains of Memory} \label{5.3}

\textbf{Overall, the increase in memory utility ($\Delta$AAUI) improves accuracy ($\Delta$Accuracy), especially under high difficulty conditions.} As shown in Figure~\ref{heatmap}, the heatmap shows that math performs best under Hard difficulty with a $\Delta$Accuracy increase of +10.67, alongside a notable increase in $\Delta$AAUI. Physics and Chemistry also show improvements, especially in Hard difficulty, where the increase in $\Delta$AAUI is closely tied to gains in $\Delta$Accuracy. In contrast, changes in $\Delta$AAUI have a smaller effect on accuracy at Easy and Medium levels.

\subsection{Inference Time Analysis} \label{5.4}
\vspace{-0.1cm}
\textbf{Overall, under the Annotated Anchors \& Attractors paradigm, the average inference time decreases by 2.1 seconds, with a performance improvement of 13.5\%.} As shown in Figure~\ref{time}, the plot demonstrates that nearly all models show varying degrees of accuracy improvement with reduced inference times when switching to the Annotated Anchors \& Attractors paradigm. Notably, larger models such as Llama-3.1-70B and Grok-4-Fast exhibit substantial performance gains and a reduction in inference time. Overall, this paradigm not only enhances model accuracy but also demonstrates a clear advantage in reducing inference latency.

\subsection{Error Type Distribution} \label{5.5}

\textbf{Overall, the models show improvements in error type distribution across the five experimental modes, particularly in Reasoning and Knowledge errors.} As shown in Figure~\ref{errorType}, the trend line chart indicates that, as the paradigms change (from Vanilla to Annotated), Reasoning and Knowledge errors decrease substantially, especially in the +Both and +Annotated paradigms, where the error rates are notably lower than in the Vanilla and +Anchors paradigms. Calculation and Formatting errors show minimal changes across the paradigms, with a slight reduction in error rates overall. Comprehension errors remain relatively stable across different paradigms, with only minor fluctuations.

\section{Related Works}
\vspace{-0.2cm}
This section reviews two strands relevant to our study: (i) memory methods for LLMs, and (ii) scientific reasoning benchmarks. 

\paragraph{Memory.}
Memory can take multiple forms, with mechanisms tailored to different needs. A common line treats memory as external, writable, and retrievable: RAG links models to non-parametric stores via indices, enabling updates at inference time~\citep{wang2024m,oche2025systematic}. Another line targets long-horizon interaction, e.g., MemGPT's ``virtual memory'' that swaps short- and long-term storage to mitigate context limits~\citep{packer2024memgptllmsoperatingsystems,wang2025karma,kang2025memory}. Agentic frameworks~\citep{chhikara2025mem0,li2025memos,zhang2026mars} store experience traces as episodic memory and improve behavior via reflection. LoCoMo~\citep{maharana2024evaluating} benchmarks long-term conversational memory.

\paragraph{Scientific Reasoning Benchmarks.}
OlympiadBench~\citep{he2024olympiadbench} provides Olympiad-level bilingual multimodal problems (notably math and physics) with expert annotations. EMMA~\citep{hao2025can} targets multimodal reasoning across math, physics, chemistry, and coding. Humanity's Last Exam (HLE)~\citep{phan2025humanity} is an expert-level, broad-coverage benchmark with a multimodal portion. MathVista~\citep{lu2023mathvista} evaluates math reasoning in visual contexts, emphasizing fine-grained perception and compositional reasoning.
ScienceBoard~\cite{sun2025scienceboard} benchmarks the scientific discovery tasks.
However, these benchmarks do not measure memory utilization during scientific reasoning.

\section{Conclusion}
\vspace{-0.2cm}

We present $A^3$-Bench, a memory-driven benchmark for scientific reasoning grounded in the activation of \emph{Anchors} and \emph{Attractors}. Consistent with human memory organization and retrieval, our design reflects hierarchical knowledge and context-dependent activation during problem solving. Using the SAPM process, we annotate 2,198 problems across math, physics, and chemistry with structured anchor units and attractor schemas. We further develop a dual-scale memory evaluation framework and the \textit{AAUI} metric to quantify memory activation during reasoning. Extensive experiments show that activation improves accuracy, keeps token consumption controllable, and exposes substantial differences in how models utilize memory. Overall, $A^3$-Bench provides a memory-centric, interpretable, and cognitively aligned evaluation paradigm that supports progress toward more human-like, memory-driven scientific reasoning.

\clearpage

\bibliography{acl_latex}

\clearpage

\appendix

\section{Free-Energy-Driven Memory Activation} \label{freeEnergy}

To better elucidate the preliminaries discussed in \S~\ref{preli}, specifically the mechanisms of anchors and attractors~\citep{zhou2025neural,siegenthaler2025visual}, the dynamics of memory activation~\citep{friston2010free}, and the paradigm of memory-augmented reasoning~\citep{ko2024memreasoner}, we formally define the memory process as ``Anchor-Induced Attractor Dynamics.'' By characterizing cognitive evolution as a trajectory within a potential energy landscape, we establish the following proposition:

\textit{\textbf{Proposition (SAAM Dynamics).}  
Let the Subject define a manifold \(\mathcal{S}\) anchored by \(\mathcal{A}\). For a reconstructed question \(q\), the memory state \(m^*\) obeys  
\(m^* = \arg\min_{m \in \mathcal{S}} \bigl[ D_{\text{KL}}(q \,\|\, p(m|\mathcal{A})) + \mathcal{H}(m) \bigr]\).  
Reasoning follows the trajectory \(\dot{m} = -\nabla \mathcal{F}\) towards stable Attractors; minimizing Free Energy thus yields Lyapunov‑like stability and facilitates memory mapping.}

\paragraph{The proof of proposition.}

We aim to prove that the memory activation state $m^*$ in a Subject domain $\mathcal{S}$ converges to a stable equilibrium under the guidance of Anchors $\mathcal{A}$ and Attractors:
\begin{equation}
m^* = \arg\min_{m \in \mathcal{S}} \mathcal{F}(m, \mathcal{A}, q),
\end{equation}
where $\mathcal{F}$ is the Variational Free Energy defined by the divergence between the question $q$ and the anchor-parameterized distribution $p(m|\mathcal{A})$.

\paragraph{Step 1: Definition of the Energy Landscape.}
Let the Subject manifold $\mathcal{S}$ be equipped with a potential energy function $U(m; \mathcal{A})$. The Anchors $\mathcal{A}$ define a set of local minima $\{a_k\}_{k=1}^N \subset \mathcal{S}$, s.t. $\nabla U(a_k) = 0$. We define the Variational Free Energy $\mathcal{F}$ as:
\begin{equation}
\mathcal{F}(m, \mathcal{A}, q) = D_{\text{KL}}(q \,\|\, p(m|\mathcal{A})) + \mathcal{H}(m),
\end{equation}
where $D_{\text{KL}}$ represents the informational "distance" (prediction error) and $\mathcal{H}$ represents the system entropy (uncertainty).

\paragraph{Step 2: Variational Minimization.}
To find the optimal memory state $m^*$, we apply the variational principle. The first-order necessary condition for a minimum at $m^*$ is the vanishing of the functional derivative:
\begin{equation}
\frac{\delta \mathcal{F}}{\delta m} \bigg|_{m^*} = 0.
\end{equation}
Expanding this, we seek a state where the drive to match the question $q$ (accuracy) is perfectly balanced by the pull of the Anchors $\mathcal{A}$ (prior knowledge) and the constraint of entropy $\mathcal{H}$.

\paragraph{Step 3: Attractor-Driven Trajectory.}
We define the reasoning process as a gradient flow in the Subject space. The temporal evolution of the mental state $\dot{m}$ follows:
\begin{equation}
\dot{m} = -\eta \nabla_m \mathcal{F}(m, \mathcal{A}, q),
\end{equation}
where $\eta$ is the learning rate or cognitive plasticity. In this dynamical system, the Anchors $\mathcal{A}$ act as **Attractors**, creating basins of attraction that pull the trajectory $\dot{m}$ toward the nearest stable fixed point.

\paragraph{Step 4: Convergence and Stability.}
By constructing $\mathcal{F}$ as a Lyapunov function, we observe that:
\begin{equation}
\frac{d\mathcal{F}}{dt} = \langle \nabla_m \mathcal{F}, \dot{m} \rangle = -\eta \|\nabla_m \mathcal{F}\|^2 \leq 0.
\end{equation}
The strictly non-positive derivative ensures that the system trajectory is dissipative and must converge to a fixed point $m^*$.

\paragraph{Conclusion.}
Thus, the reasoning process:
\begin{equation}
\hat{m} = \arg\min_{m} \left[ D_{\text{KL}}(q \,\|\, p(m|\mathcal{A})) + \mathcal{H}(m) \right]
\end{equation}
formally links the external query $q$ with internal anchor-based structures, ensuring that memory activation is an emergent property of Attractor Dynamics and Free Energy Minimization.

\section{Annotation Guidance} \label{annotationGuidance}

This section presents the manual annotation guidelines for each step in \S~\ref{section3}.

\subsection{Anchor \& Attractor Developing}

\begin{itemize}
    \item \textbf{Subdomain Definitions and Frameworks:}
    Before annotation, experts should delineate each subdomain's scope and key reasoning framework, so that the boundaries and core concepts are explicit. This provides a stable reference for selecting anchors and attractors without drifting across subdomains.

    \item \textbf{Anchors Identification:}
    Anchors are the foundational reasoning primitives (e.g., core concepts, theorems, formulas). Experts should extract representative anchors from the subdomain's canonical content, prioritizing items that are broadly reusable and frequently invoked as starting principles in problem solving.

    \item \textbf{Attractors Identification:}
    Attractors connect abstract principles to actionable solution pathways. Each attractor contains an \textit{Abstract Schema} (a reusable solution template grounded in anchors) and \textit{Episodic Exemplars} (concrete instantiations). Experts should curate attractors that reliably operationalize anchors for typical scientific tasks.

    \item \textbf{Anchor--Attractor Relations and Consistency:}
    After extraction, experts should specify which anchors support each attractor and how they interact during reasoning, ensuring the mapping is coherent, non-overlapping, and free of redundancy. The role of each unit should be unambiguous in the reasoning chain.

    \item \textbf{Library Construction:}
    Finally, all anchors and attractors are organized into an \textit{Anchor Library} and an \textit{Attractor Library} in \textit{JSON} format. Each entry includes an identifier, a concise definition, and explicit relations to enable scalable management, retrieval, and traceability.
\end{itemize}

\subsection{Problem Reconstructing}

\begin{itemize}
    \item \textbf{Error Diagnosis:}
    Experts review the original question alongside LLM answers to identify common failure modes (e.g., missing steps, wrong assumptions, incomplete derivations), analyze their causes, and revise the question accordingly.

    \item \textbf{Cross-Model Refinement:}
    For second-stage questions, experts compare outputs from three LLMs to surface systematic discrepancies and uncovered knowledge gaps, then refine the problem to elicit essential reasoning steps and broader scientific coverage.

    \item \textbf{Multi-hop Enforcement:}
    Reconstructed questions require at least two coupled knowledge points with an explicit stepwise dependency, preventing one-shot solutions and promoting multi-stage reasoning.

    \item \textbf{Reference Solution Writing:}
    Each revised problem is paired with a correct, complete standard answer that includes the final result and the key intermediate steps.
\end{itemize}

\subsection{Memory Mapping}

\begin{itemize}
    \item \textbf{Subdomain Assignment:}
    For each question, three LLMs propose candidate subdomains and a voting rule selects the final label; experts then verify the assignment and confirm the secondary discipline to ensure accurate classification.

    \item \textbf{Anchor--Attractor Annotation:}
    Given the confirmed discipline, experts manually annotate \emph{Anchors} (e.g., core concepts, theorems, formulas) and \emph{Attractors} (e.g., abstract schemas with episodic exemplars) following the annotation principles, where anchors support key reasoning steps and attractors operationalize them into solution pathways.

    \item \textbf{Quantity Control and Consistency:}
    Each question is capped at no more than 6 annotated units in total (anchors + attractors) to keep complexity moderate; experts also check that selected items are necessary, non-redundant, and tightly aligned with the intended reasoning process.

    \item \textbf{Rationale Logging:}
    Experts record brief justifications for the chosen anchors and attractors and summarize the annotation process per question, enabling traceability and efficient future review.
\end{itemize}

\section{Subject Taxonomy}
\label{datasetTaxonomy}

As shown in Table~\ref{subdomains}, this section outlines the dataset taxonomy across three subject domains: \textit{Math}, \textit{Physics}, and \textit{Chemistry}. For each domain, we first state the primary classification criteria and then introduce the corresponding second-level subdomains. This design emphasizes the hierarchical organization of disciplinary knowledge and clarifies the distinctions across domains.

\subsection{Math}

The math subset follows the \textit{Mathematics Subject Classification (MSC)} system~\citep{zbmath_msc2020, arndt2021msc, msc_digital_world}, jointly developed and maintained by the American Mathematical Society (AMS) and zbMATH Open~\citep{msc2020_report}. As a globally adopted indexing scheme for journals, scholarly databases, and university curricula, MSC provides a stable and internationally comparable foundation for benchmarking LLMs~\citep{huang2025chartsketcher,wucausal}. Building on MSC primary classes (00--99) and modern mathematical organization, we group problems into eight subdomains: \textit{Algebra}, \textit{Geometry}, \textit{Number Theory}, \textit{Mathematical Analysis}, \textit{Discrete Math}, \textit{Logic \& Set Theory}, \textit{Statistics \& Decision Sciences}, and \textit{Computational Math}.

In brief, these subdomains cover core algebraic structures, spatial reasoning and invariants, integer arithmetic and congruences, limits/calculus and infinite processes, combinatorics and graph structures, formal foundations, probabilistic inference under uncertainty, and numerical/scientific computing. This taxonomy enables systematic assessment of mathematically grounded retrieval, abstraction across concept hierarchies, and domain-specific reasoning strategies.

\begin{table}[t]
    \centering
    \small
    \setlength{\tabcolsep}{8pt}
    \renewcommand{\arraystretch}{1.1}
    \resizebox{\columnwidth}{!}{
    \begin{tabular}{l l r}
        \toprule
        \textbf{Subject} & \textbf{Subdomain} & \textbf{Count} \\
        \midrule
        \multirow{8}{*}{\textbf{Math}}
            & \hspace{1em} Algebra                 & 158 \\
            & \hspace{1em} Geometry                & 120 \\
            & \hspace{1em} Number Theory           & 108 \\
            & \hspace{1em} Calculus \& Analysis    & 132 \\
            & \hspace{1em} Discrete Math           & 132 \\
            & \hspace{1em} Logic \& Set Theory     & 96 \\
            & \hspace{1em} Statistics \& Probability     & 120 \\
            & \hspace{1em} Computational Math      & 132 \\
        \midrule
        \multirow{5}{*}{\textbf{Physics}}
            & \hspace{1em} Mechanics               & 120 \\
            & \hspace{1em} Thermodynamics          & 120 \\
            & \hspace{1em} Optics                  & 120 \\
            & \hspace{1em} Electromagnetism        & 120 \\
            & \hspace{1em} Modern Physics          & 120 \\
        \midrule
        \multirow{5}{*}{\textbf{Chemistry}}
            & \hspace{1em} Inorganic Chemistry     & 120 \\
            & \hspace{1em} Organic Chemistry       & 120 \\
            & \hspace{1em} Physical Chemistry      & 120 \\
            & \hspace{1em} Analytical Chemistry    & 120 \\
            & \hspace{1em} Biochemistry            & 120 \\
              \midrule
        \textbf{Grand Total} & & \textbf{2,198} \\
        \bottomrule
    \end{tabular}}
    \caption{Subdomain composition of the $A^3$-Bench.}
    \label{subdomains}
\end{table}

\subsection{Physics}
The Physics taxonomy follows standard higher-education curricula and internationally recognized physics classification standards~\citep{smith2020physics}, aligned with prior work on broad LLM evaluation and alignment~\citep{hendryckstest2021, hendrycks2021ethics} and recent benchmarks for complex scientific reasoning and agentic exploration~\citep{he2024olympiadbench, li2023camel}. We organize the dataset into five canonical subdomains: \textit{Mechanics}, \textit{Thermodynamics}, \textit{Optics}, \textit{Electromagnetism}, and \textit{Modern Physics}. This division supports systematic evaluation of whether models can retrieve appropriate physical principles (anchors) and apply scenario-specific solution patterns (attractors) across abstraction levels.

Mechanics covers kinematics, dynamics, and conservation laws, emphasizing force analysis and motion under constraints. Thermodynamics focuses on heat, work, and state variables, including entropy-driven reasoning and statistical-mechanical interpretations. Optics spans geometric optics (imaging, lenses, ray tracing) and wave optics (interference, diffraction, polarization), requiring careful treatment of limiting assumptions. Electromagnetism addresses charges, fields, and circuits, involving vector-field reasoning and circuit-level modeling. Modern Physics extends beyond classical theory to relativity and quantum phenomena, where counter-intuitive effects demand strict adherence to formal principles.

\subsection{Chemistry}

The Chemistry taxonomy follows the canonical organization of chemical science education and advanced evaluation frameworks~\citep{hendryckstest2021, hendrycks2021ethics}, aligned with internationally recognized IUPAC standards~\citep{heller2013inchi} and recent benchmarks for chemical intelligence~\citep{runcie2025assessing, li2023camel, feng2024sciknoweval, zhao2025superchemmultimodalreasoningbenchmark}. We group problems into five subdomains: \textit{Inorganic Chemistry}, \textit{Organic Chemistry}, \textit{Physical Chemistry}, \textit{Analytical Chemistry}, and \textit{Biochemistry}, enabling fine-grained evaluation of knowledge retrieval and structured reasoning across chemical contexts.

Inorganic Chemistry covers the synthesis, structure, and reactivity of inorganic and organometallic compounds. Organic Chemistry focuses on carbon-based molecules, emphasizing functional groups, reaction mechanisms, and synthesis planning. Physical Chemistry studies chemical systems through thermodynamics, kinetics, and quantum principles. Analytical Chemistry concerns qualitative and quantitative determination via methods such as spectroscopy and chromatography. Biochemistry examines biomolecular structure and function, linking chemical mechanisms to metabolic pathways and cellular processes.

\begin{table*}[htb]
\centering
\setlength{\tabcolsep}{3pt}
\resizebox{\linewidth}{!}{
\begin{tabular}{l|cccc|cccc|cccc|ccc}
\toprule
\multirow{2}{*}{\textbf{Method}} &
\multicolumn{4}{c|}{\textbf{Math}} &
\multicolumn{4}{c|}{\textbf{Physics}} &
\multicolumn{4}{c|}{\textbf{Chemistry}} &
\multirow{2}{*}{\textbf{Avg.}} &
\multirow{2}{*}{\textbf{AAUI}} &
\multirow{2}{*}{\textbf{Tokens}} \\
& \textbf{Easy} & \textbf{Medium} & \textbf{Hard} & \textbf{Avg.}
& \textbf{Easy} & \textbf{Medium} & \textbf{Hard} & \textbf{Avg.}
& \textbf{Easy} & \textbf{Medium} & \textbf{Hard} & \textbf{Avg.}
& & & \\
\midrule

\rowcolor{lightgray}
\multicolumn{16}{c}{\textbf{Vanilla + Anchor-Only}} \\
\midrule
DeepSeek-V3.2 & 60.90 & 49.16 & 28.00 & 47.49 & \textbf{64.58} & \textbf{58.33} & \textbf{43.89} & \textbf{56.50} & \underline{69.58} & \underline{56.67} & 37.22 & \underline{56.00} & \textbf{52.27} & 0.44 & $1.07\times10^{6}$ \\
Gemini-2.5-Flash & 31.33 & 29.10 & 26.00 & 29.06 & 2.50 & 2.78 & 5.00 & 3.33 & 5.42 & 5.00 & 10.56 & 6.83 & 15.97 & 0.08 & $1.71\times10^{6}$ \\
Claude-Haiku-4.5 & \underline{62.66} & \underline{52.51} & 30.67 & \underline{50.00} & 57.92 & 45.00 & 29.44 & 45.50 & \textbf{70.00} & \textbf{57.22} & 37.22 & \textbf{56.33} & 50.50 & 0.43 & $1.49\times10^{6}$ \\
Grok-4-Fast & \textbf{63.91} & \textbf{53.51} & \textbf{36.67} & \textbf{52.61} & \underline{60.42} & \underline{56.67} & 30.56 & \underline{50.33} & 67.50 & 49.44 & \underline{38.33} & 53.33 & \underline{52.18} & 0.47 & $1.80\times10^{6}$ \\
GPT-5-Mini & 31.08 & 27.76 & 29.00 & 29.46 & 13.75 & 10.56 & 14.44 & 13.00 & 23.75 & 22.78 & 18.89 & 22.00 & 22.93 & 0.26 & $1.73\times10^{6}$ \\
Qwen3-4B & 51.63 & 51.51 & \underline{36.67} & 47.09 & 39.17 & 31.67 & 20.00 & 31.17 & 54.58 & 47.78 & 35.56 & 46.83 & 42.68 & 0.33 & $1.87\times10^{6}$ \\
Qwen3-30B & 53.38 & 48.16 & 33.67 & 45.89 & 52.50 & 36.67 & 28.33 & 40.50 & 47.50 & 47.78 & 27.78 & 41.67 & 43.27 & 0.38 & $1.83\times10^{6}$ \\
Llama-3.1-70B & 47.12 & 39.80 & 26.67 & 38.78 & 37.50 & 28.89 & 17.78 & 29.00 & 48.33 & 41.11 & 28.33 & 40.17 & 36.49 & 0.29 & $1.22\times10^{6}$ \\
GLM-4-32B & 57.89 & 45.48 & 28.33 & 45.29 & 39.58 & 33.33 & 18.89 & 31.50 & 41.67 & 34.44 & 21.11 & 33.33 & 38.26 & 0.36 & $9.33\times10^{5}$ \\
GPT-OSS-120B & 55.64 & 47.49 & 34.67 & 46.89 & 50.00 & 38.33 & \underline{32.78} & 41.33 & 55.83 & 52.78 & \textbf{47.22} & 52.33 & 46.86 & 0.42 & $1.45\times10^{6}$ \\
\midrule

\rowcolor{lightgray}
\multicolumn{16}{c}{\textbf{Vanilla + Attractor-Only}} \\
\midrule
DeepSeek-V3.2 & 62.16 & \underline{57.19} & 30.33 & 51.10 & \textbf{74.17} & \underline{53.89} & \underline{38.33} & \underline{57.33} & \underline{70.83} & \textbf{65.00} & \underline{43.33} & \textbf{60.83} & \underline{55.46} & 0.46 & $1.24\times10^{6}$ \\
Gemini-2.5-Flash & 35.84 & 36.79 & 24.33 & 32.67 & 2.92 & 4.44 & 3.89 & 3.67 & 8.33 & 10.00 & 12.78 & 10.17 & 18.61 & 0.22 & $1.94\times10^{6}$ \\
Claude-Haiku-4.5 & \underline{66.42} & \textbf{59.20} & 30.33 & \underline{53.41} & 61.67 & 48.33 & 30.00 & 48.17 & \textbf{72.92} & \underline{63.33} & 40.56 & \underline{60.33} & 53.87 & 0.31 & $1.74\times10^{6}$ \\
Grok-4-Fast & \textbf{72.93} & 55.85 & \textbf{42.33} & \textbf{58.62} & \underline{70.42} & \textbf{57.78} & \textbf{43.89} & \textbf{58.67} & 67.92 & 53.33 & 38.89 & 54.83 & \textbf{57.60} & 0.49 & $2.17\times10^{6}$ \\
GPT-5-Mini & 26.82 & 26.42 & 28.33 & 27.15 & 18.75 & 12.22 & 17.22 & 16.33 & 20.00 & 22.22 & 18.89 & 20.33 & 22.34 & 0.36 & $1.95\times10^{6}$ \\
Qwen3-4B & 47.37 & 45.15 & 36.33 & 43.39 & 32.08 & 24.44 & 17.22 & 25.33 & 50.00 & 50.00 & 33.33 & 45.00 & 38.90 & 0.40 & $1.88\times10^{6}$ \\
Qwen3-30B & 58.15 & 51.17 & \underline{38.67} & 50.20 & 58.33 & 38.89 & 34.44 & 45.33 & 58.33 & 51.67 & 30.56 & 48.00 & 48.27 & 0.42 & $1.85\times10^{6}$ \\
Llama-3.1-70B & 54.64 & 44.15 & 29.67 & 43.99 & 43.75 & 30.00 & 21.11 & 32.83 & 55.00 & 47.22 & 32.22 & 45.83 & 41.45 & 0.42 & $1.24\times10^{6}$ \\
GLM-4-32B & 62.66 & 54.52 & 31.00 & 50.70 & 47.92 & 32.78 & 26.11 & 36.83 & 48.33 & 43.89 & 26.11 & 40.33 & 44.09 & 0.40 & $1.19\times10^{6}$ \\
GPT-OSS-120B & 56.64 & 46.15 & 30.33 & 45.59 & 47.08 & 33.89 & 27.22 & 37.17 & 57.08 & 56.67 & \textbf{48.33} & 54.33 & 45.68 & 0.41 & $1.66\times10^{6}$ \\
\bottomrule
\end{tabular}}
\caption{Performance comparison of ten LLMs under two partial activation paradigms: vanilla + anchor-only versus vanilla + attractor-only.}
\label{tab:other_results}
\end{table*}

\newpage

\section{Different Memory Paradigms}
\label{differnetParadigms}

To isolate the contributions of Anchors and Attractors, we conduct two different paradigms that activate only anchors or only attractors. Table~\ref{tab:other_results} reports the results, leading to four findings.

\paragraph{Attractors tend to contribute more than Anchors when only one memory type is available.}
For most models (7/10), Attractor-only activation yields higher overall accuracy than Anchor-only activation.
For example, Grok-4-Fast improves from 52.18\% (Anchor-only) to 57.60\% (Attractor-only), and GLM-4-32B increases from 38.26\% to 44.09\%.
Similar trends hold for Llama-3.1-70B (36.49\%$\rightarrow$41.45\%) and Qwen3-30B (43.27\%$\rightarrow$48.27\%).
This suggests that, under partial memory, access to procedural schemas is often more directly useful for completing multi-step scientific derivations.

\paragraph{Both memory types are needed to reach the best performance.}
Despite the strength of Attractor-only activation, it remains consistently below the dual annotated paradigm.
For example, Grok-4-Fast increases from 57.60\% (Attractor-only) to 65.10\% (Annotated dual), and Qwen3-30B increases from 48.27\% to 60.60\%.
The gap is especially large for Qwen3-4B (38.90\%$\rightarrow$58.92\%), indicating that procedural templates alone are insufficient without the supporting conceptual grounding.
Overall, Anchors and Attractors play complementary roles: templates provide the solution pathway, while definitions and constraints help instantiate the pathway correctly for the specific problem.

\paragraph{Subjects show different sensitivities to memory types.}
The relative advantage of Attractors over Anchors varies by domain.
In Chemistry, the difference is often small (e.g., Grok-4-Fast: 53.33\% vs.\ 54.83\%; Claude-Haiku-4.5: 56.33\% vs.\ 60.33\%), consistent with the need for precise property definitions alongside procedures.
In contrast, Math and Physics more frequently favor Attractor-only activation, reflecting the procedural nature of theorem selection and template-based derivations (e.g., Grok-4-Fast: Math 52.61\%$\rightarrow$58.62\%, Physics 50.33\%$\rightarrow$58.67\%).

\paragraph{Dependence on complete memory support differs substantially across models.}
Removing either component can cause large drops for some models but only modest changes for others.
For instance, Qwen3-4B falls from 58.92\% (Annotated dual) to 42.68\% (Anchor-only) and 38.90\% (Attractor-only), and Qwen3-30B drops from 60.60\% to 43.27\% and 48.27\%.
By contrast, GPT-OSS-120B changes only slightly (47.18\% to 46.86\% and 45.68\%).
These results indicate that robustness to partial memory is model-dependent, and that the ability to integrate both Anchors and Attractors is a key limitation exposed by the dataset.

\begin{table}[t]
    \centering
    \small
    \resizebox{\columnwidth}{!}{
    \begin{tabular}{lcc}
        \toprule
        \textbf{Comparison} & \textbf{$p$-value} & \textbf{Significance} \\
        \midrule
        Q vs. A            & $< 0.001$  & \checkmark \\
        Q vs. T            & $< 0.001$  & \checkmark \\
        A vs. T            & $0.120$    & $\times$   \\
        A vs. AT           & $< 0.001$  & \checkmark \\
        T vs. AT           & $< 0.001$  & \checkmark \\
        AT vs. AT$^*$      & $< 0.001$  & \checkmark \\
        \bottomrule
    \end{tabular}}
    \caption{Results of McNemar’s Test for statistical significance. Q: question only; A: +anchors; T: +attractors; AT: +both; AT$^*$: +annotated . Significance level $\alpha = 0.05$.}
    \label{tab:significance_test}
\end{table}

\section{Other Analysis}
\label{otherAnalysis}
We conduct additional experiments, including a task-appropriate McNemar’s test for statistical significance and an analysis of robustness under noisy memory interference.

\subsection{Significance Test}

To quantitatively assess the necessity of memory augmentation for scientific reasoning, we apply McNemar’s test~\citep{pembury2020effective} to the prediction outcomes. As summarized in Table~\ref{tab:significance_test}, the results provide statistical evidence for the reliance of scientific reasoning on external context.

First, the comparison between the question-only baseline and single-memory activation ($Q$ vs.\ $A/T$, $p<0.001$) shows that parametric knowledge alone is insufficient for solving complex scientific problems. Introducing external memory, whether declarative or procedural, leads to a statistically significant improvement in performance.

Second, the consistent advantage of combined activation ($AT$) over either component alone ($p<0.001$) indicates that scientific reasoning operates as a dual-process mechanism. Anchors and Attractors each contribute essential information, and effective reasoning requires their joint activation rather than isolated use.

Finally, the absence of a significant difference between Anchor-only and Attractor-only activation ($p=0.120$) suggests that the two memory types play complementary and comparably important roles. Neither dominates the other; instead, they function as parallel cognitive supports whose integration yields the strongest reasoning performance.

\begin{figure}[t]
	\large
	\centering
	\includegraphics[width=0.8\columnwidth]{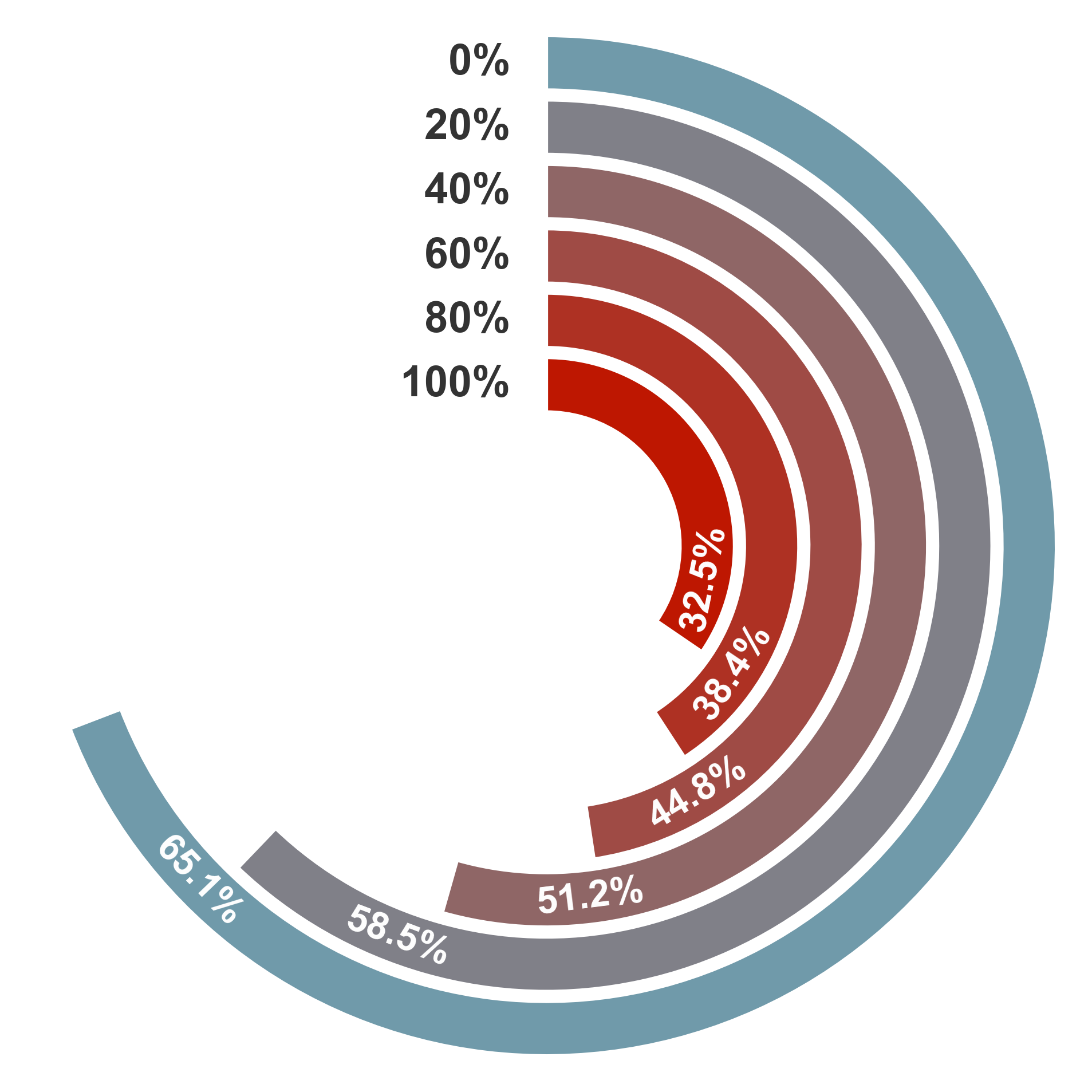}
	\caption{Impact of memory relevance vs. Noise on model performance.This radial bar chart illustrates the degradation of accuracy in Grok-4-Fast as high-quality annotated memory (anchors \& attractors) is progressively replaced by irrelevant noise memory. The concentric rings correspond to increasing noise replacement ratios (from 0\% to 100\%).}
	\label{noise}
\end{figure}

\subsection{Noise Interference}

As illustrated in Figure~\ref{noise}, increasing the noise replacement ratio consistently reduces model accuracy, highlighting that performance depends not only on \emph{having} memory but on the \emph{relevance} of the activated memory.

For Grok-4-Fast, accuracy declines monotonically from 65.1\% with 100\% annotated memory to 58.5\% (20\% noise), 51.2\% (40\%), 44.8\% (60\%), 38.4\% (80\%), and 32.5\% under full noise replacement. Notably, the drop becomes pronounced once noise exceeds 40--60\%, suggesting that irrelevant memory increasingly dominates the context, distracts the model from key principles, and disrupts the anchor--attractor alignment needed to initiate correct solution paths. The final performance at 100\% noise approaches a near-memoryless regime, indicating that low-quality memory can effectively negate the benefits of retrieval and even harm reasoning by introducing misleading cues.

\begin{figure*}[t]
    \centering
    \includegraphics[width=0.85\textwidth]{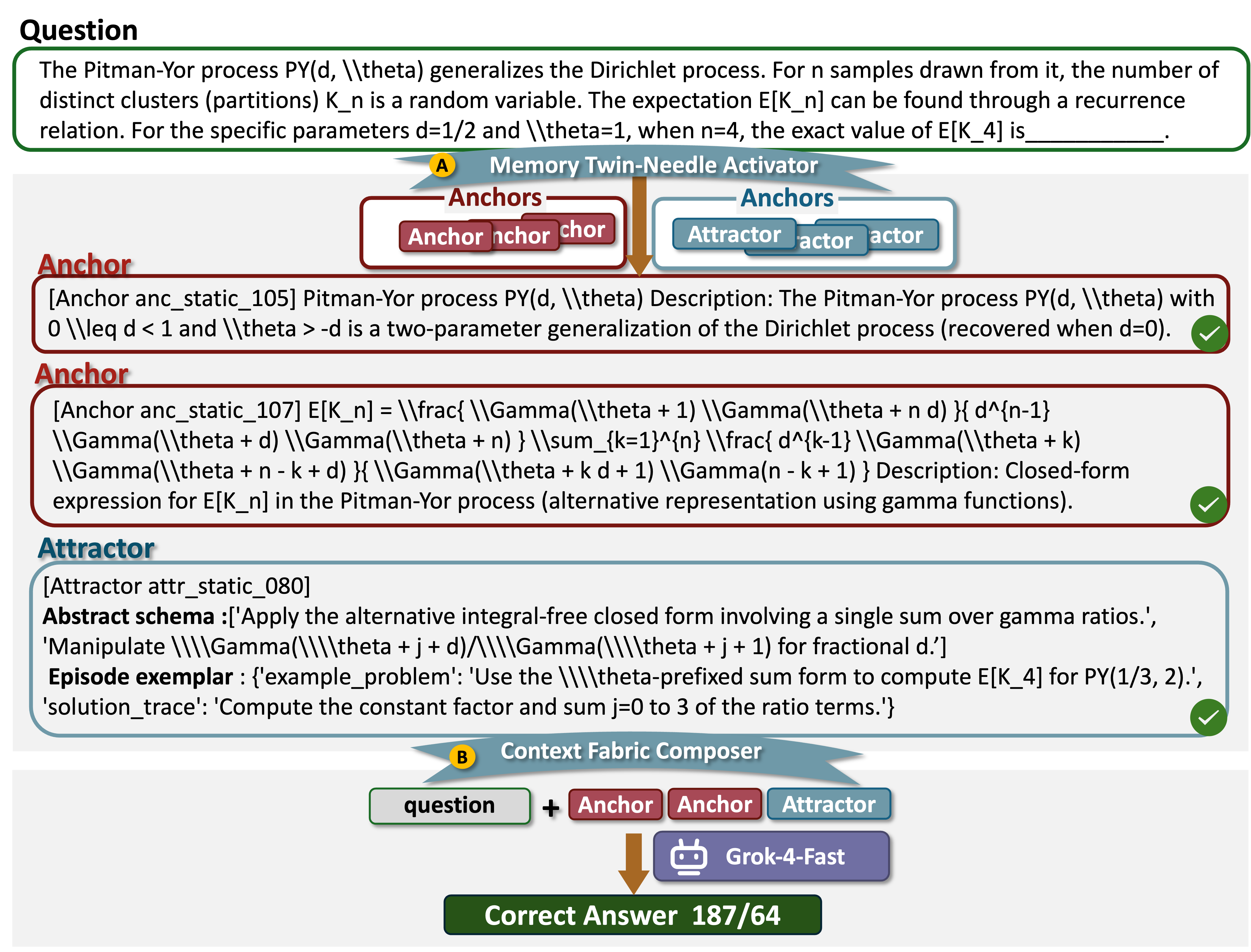}
    \caption{Successful case of Grok-4-Fast on a TheoremQA problem under anchor \& attractor activation. Using HybridRAG, the model activates relevant anchors and attractors from the full memory repositories, composes them with the question, and produces the correct answer (187/64).}
    \label{cor_case}

    \vspace{0.8em}

    \includegraphics[width=0.85\textwidth]{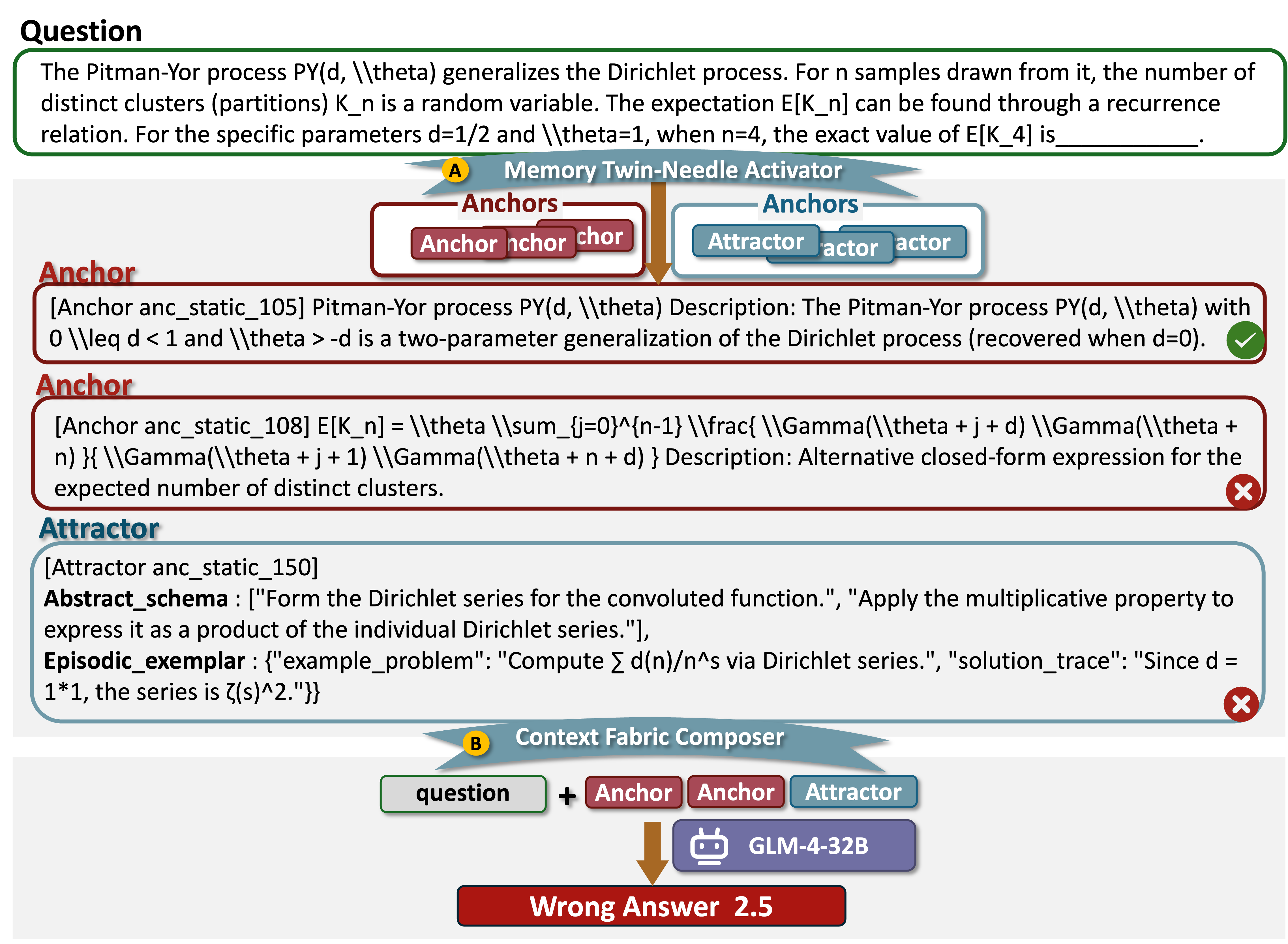}
    \caption{Failure case of GLM-4-32B on a TheoremQA problem under anchor \& attractor activation. Although HybridRAG is used, the retrieved anchors/attractors are irrelevant and provide little support; after composition with the question, the model outputs an incorrect answer (2.5).}
    \label{error_case}
\end{figure*}

\section{Case Study}
\label{caseStudy}

To illustrate how memory activation influences reasoning outcomes, we present two representative cases involving the Pitman--Yor process, shown in Figures~\ref{cor_case} and~\ref{error_case}.

\paragraph{Successful activation (Figure~\ref{cor_case}).}
In the successful case, the Memory Twin-Needle Activator retrieves both the core definition of the Pitman--Yor process (Anchor \texttt{anc\_105}) and the relevant closed-form expectation formula (Anchor \texttt{anc\_107}). Importantly, it also activates the correct Attractor (\texttt{attr\_080}), which encodes an abstract reasoning schema for manipulating Gamma-function ratios, along with an episodic exemplar for computing $E[K_n]$.
With access to both the conceptual grounding (definitions) and the procedural guidance (schema), the model (Grok-4-Fast) successfully completes the multi-step derivation and obtains the correct result, $187/64$.
This case demonstrates that Attractors function as procedural guides, enabling reasoning trajectories that would be correct and efficiency.

\paragraph{Failure activation (Figure~\ref{error_case}).}
The failure case highlights the fragility of reasoning under imprecise memory activation. Although the correct subject definition is retrieved (\texttt{anc\_105}), the activation result contains substantial noise. The Vector Needle selects a distracting Anchor (\texttt{anc\_108}), a complex closed-form expression that shifts the reasoning away from the intended recurrence-based approach. Meanwhile, the Graph Needle activates an irrelevant Attractor (\texttt{attr\_150}) associated with Dirichlet series and multiplicative number theory.

As a consequence, the model (GLM-4-32B) follows an incompatible procedure and fails to produce a valid derivation, yielding an incorrect value ($2.5$). This example shows that identifying the correct topic alone is insufficient. Reliable reasoning requires coherent alignment between declarative content (Anchors) and procedural guidance (Attractors).

\end{document}